\documentclass[journal]{IEEEtran}
\usepackage{cite}

\ifCLASSINFOpdf
\else
\fi
\usepackage{bbding}
\usepackage{times}
\usepackage{epsfig}
\usepackage{graphicx}
\usepackage{caption}
\graphicspath{{figures/}}
\usepackage{amsmath}
\usepackage{amssymb}
\usepackage[colorlinks,linkcolor=red]{hyperref}

\usepackage[utf8]{inputenc} 
\usepackage[T1]{fontenc}    
\usepackage{url}            
\usepackage{booktabs}       
\usepackage{amsfonts}       
\usepackage{nicefrac}       
\usepackage{microtype}      
\usepackage{array}
\usepackage{bm}
\usepackage{multirow}

\usepackage{times} 
\usepackage{helvet} 
\usepackage{courier} 
\usepackage{amsmath}
\usepackage{footnote}
\usepackage{multirow}
\usepackage{array}
\usepackage{algorithm}
\usepackage{algpseudocode}
\usepackage{amsfonts,amssymb}
\usepackage{booktabs}
\usepackage{color}
\usepackage{verbatim}

\begin{document}
	%
	\title{CDE-GAN: Cooperative Dual Evolution Based Generative Adversarial Network}
	%
	%
	%
	
	\author{
		Shiming~Chen,
		Wenjie~Wang,
		Beihao~Xia,
		Xinge~You,~\IEEEmembership{Senior~Member,~IEEE},
		Qinmu~Peng,
		Zehong~Cao,
		and~Weiping~Ding,~\IEEEmembership{Senior~Member,~IEEE}
		
		\thanks{ This work was supported in part by the National Natural Science Foundation of China~(61571205, 61772220 and 61976120), the Key Program for International S\&T Cooperation Projects of China~(2016YFE0121200), the Special Projects for Technology Innovation of Hubei Province~(2018ACA135), the Key Science and Technology Innovation Program of Hubei Province~(2017AAA017),  the Natural Science Foundation of Jiangsu Province BK~(20191445), the Natural Science Foundation of Hubei Province~(2018CFB691), fund from Science, Technology and Innovation Commission of Shenzhen Municipality~(JCYJ20180305180804836 and JSGG20180507182030600).(Corresponding author: Xinge You.)}
		\thanks{S. Chen, W. Wang, Beihao Xia are with the School of Electronic Information and Communication, Huazhong University of Science and Technology, Wuhan 430074, China. (e-mail:gchenshiming@gmail.com; shimingchen@hust.edu.cn)}
		\thanks{Qinmu Peng and Xinge You are with the Shenzhen Research Institute, Huazhong University of Science and Technology, Shenzhen 518000, China, and also with the School of Electronic Information and Communications, Huazhong University of Science and Technology, Wuhan 430074, China. (e-mail:youxg@hust.edu.cn)}
		\thanks{Z. Cao is is with the Discipline of ICT, University of Tasmania, TAS 7001, Australia.}
		\thanks{W. Ding is with the School of Information Science and Technology, Nantong University, Nantong, China.}
	}

	\markboth{IEEE Transactions on Evolutionary Computation}%
	{Shell \MakeLowercase{\textit{et al.}}: Bare Demo of IEEEtran.cls for IEEE Journals}
	%



	\maketitle
	
	\begin{abstract}
		Generative adversarial networks (GANs) have been a popular deep generative model for real-world applications.  Despite many recent efforts on GANs that have been contributed,  mode collapse and instability of GANs are still open problems caused by their adversarial optimization difficulties. In this paper, motivated by the cooperative co-evolutionary algorithm, we propose a \textit{Cooperative Dual Evolution based Generative Adversarial Network} (CDE-GAN) to circumvent these drawbacks. In essence, CDE-GAN  incorporates dual evolution with respect to the generator(s) and discriminators into a unified evolutionary adversarial framework to conduct effective adversarial multi-objective optimization. Thus it exploits the complementary properties and injects dual mutation diversity into training to steadily diversify the estimated density in capturing multi-modes and improve generative performance. Specifically, CDE-GAN decomposes the complex adversarial optimization problem into two subproblems (generation and discrimination), and each subproblem is solved with a separated subpopulation (\textit{E-Generators} and \textit{E-Discriminators}), evolved by its own evolutionary algorithm. Additionally, we further propose a \textit{Soft Mechanism} to balance the trade-off between E-Generators and E-Discriminators to conduct steady training for CDE-GAN. Extensive experiments on one synthetic dataset and three real-world benchmark image datasets demonstrate that the proposed CDE-GAN achieves a competitive and superior performance in generating good quality and diverse samples over baselines. The code and more generated results are available at our project homepage \textit{\url{https://shiming-chen.github.io/CDE-GAN-website/CDE-GAN.html}}.
	\end{abstract}
	

	%
	\IEEEpeerreviewmaketitle

	\section{Introduction and Motivation}\label{sec1}
	%
	%
	%
	%
	\IEEEPARstart{G}{enerative} adversarial networks (GANs) \cite{Goodfellow2014GenerativeAN} are new popular methods for generative modeling, using game-theoretic training schemes to implicitly learn a given probability density. With the potential power of capturing high dimensional probability distributions, GANs have been successfully deployed for various synthesis tasks, e.g., image generation \cite{Zhu2017UnpairedIT}, video prediction \cite{tulyakov2018mocogan,Xie2020CooperativeTO}, text synthesis \cite{Yu2017SeqGANSG,Liu2020CatGANCG}.

	A GAN consists of a framework describing the interaction between two different models, i.e., generator and discriminator, which are used to solve a min-max game optimization problem. During learning, the generator tries to learn real data distribution by generating realistic-looking samples that can fool the discriminator, while the discriminator attempts to differentiate between samples from the data distribution and the ones produced by the generator. Although the pioneered GAN provided some analysis on the convergence properties of the approach \cite{Goodfellow2014GenerativeAN,Goodfellow2016NIPS2T}, it assumed that updates occurred in pure function space, allowed arbitrarily powerful generator and discriminator networks, and modeled the resulting optimization objective as a convex-concave gam. Therefore it is yielding well-defined global convergence properties. Besides, this analysis assumed that the discriminator network is fully optimized between generator updates. However, these assumptions do not mirror the practice of GAN optimization. In pratice, there exist many well-documented failure GAN models caused by mode collapse\footnote{the generator can only learn some limited patterns from the large-scale given datasets, or assigns all of its probability mass to a small region in the space \cite{Arora2017GeneralizationAE}} or instability\footnote{the discriminator can easily distinguish between real and fake samples during the training phase\cite{Neyshabur2017StabilizingGT,Berthelot2017BEGANBE,Arjovsky2017WassersteinGA}}. 
	
	Therefore, many recent efforts on GANs have contributed to overcome these optimization difficulties by developing various adversarial training approaches, i.e., modifying the optimization objective, training additional discriminators, training multiple generators, and using evolutionary computation. The first method is a typical way to control the optimization gradient of discriminator or generator parameters, which would help GANs to steadily arrive at the equilibrium points of the optimization procedure under proper conditions \cite{Arjovsky2017WassersteinGA,Metz2017UnrolledGA,WardeFarley2017ImprovingGA,Gulrajani2017ImprovedTO,Miyato2018SpectralNF}. Compared to the traditional GANs performed single-objective optimization, the second method revisits the multiple-discriminator setting by framing the simultaneous optimization of different discriminator models as a multi-objective optimization problem. Thus it would overcome the problem involving the lack of informative gradient (i.e., the stable gradient is neither vanishing nor exploding for the generator throughout training \cite{Neyshabur2017StabilizingGT,Gulrajani2017ImprovedTO}.) signal provided by discriminator \cite{Nguyen2017DualDG,Durugkar2017GenerativeMN,Neyshabur2017StabilizingGT,Doan2019OnlineAC,Albuquerque2019MultiobjectiveTO}. The third method simultaneously trains multiple generators with the object that mixture of their induced distributions would approximate the data distribution, and thus mode collapse problem can be alleviated \cite{Tolstikhin2017AdaGANBG,Arora2017GeneralizationAE,Hoang2018MGANTG,Ghosh2018MultiagentDG}. However, the aforementioned GAN methods are limited by the specified adversarial optimization strategy. The last method introduced evolutionary computation for the optimization of GANs to improve generative performance \cite{Schmiedlechner2018TowardsDC,Wang2019EvolutionaryGA,Costa2019COEGANET,Toutouh2019SpatialEG,Liu2020CatGANCG}. In fact, the existing evolutionary GANs evolve a population of generators (e.g., E-GAN \cite{Wang2019EvolutionaryGA}, CatGAN \cite{Liu2020CatGANCG}) or GANs (Lipizzar \cite{Schmiedlechner2018TowardsDC}, Mustangs \cite{Toutouh2019SpatialEG}) to play the adversarial game, which will result that GANs evolve in a static environment\footnote{when evolving the generators of GAN, its discriminator acts as the environment, e.g., \cite{Wang2019EvolutionaryGA}; and vice versa.}. In essential, training GANs is a large-scale optimization problem, which is challenging \cite{Wang2019EvolutionaryGA,Albuquerque2019MultiobjectiveTO}. This is why the most existing GAN methods usually fall in instability and mode collapse. To overcome these challenges, we propose a novel insight for GANs, in which a cooperative dual evolution paradigm is proposed to conduct adversarial multi-objective optimization for GANs.
	
	Recently, evolutionary computation (EC) has been used to solve many deep learning challenges, e.g., optimizing deep learning hyper-parameters \cite{Sun2019EvolvingUD,Sun2019APS,Sun2020EvolvingDC,Sun2020SurrogateAssistedED,Toutouh2020RepurposingHG} and designing network architecture \cite{Sun2020CompletelyAC,Sun2020AutomaticallyDC}. Meanwhile, researchers also attempt to apply EC on GANs to improve the robust of GANs for mitigating degenerate GAN dynamics \cite{Schmiedlechner2018TowardsDC,Wang2019EvolutionaryGA,Costa2019COEGANET,Toutouh2019SpatialEG,Liu2020CatGANCG}. Among them, Lipizzaner \cite{Schmiedlechner2018TowardsDC} and Mustangs \cite{Toutouh2019SpatialEG} use a spatial distributed competitive co-evolutionary algorithm to provide diversity in the genome space for GANs. In \cite{Wang2019EvolutionaryGA}, Wang introduced E-GAN that injects diversity into the training with the evolutionary population. Accordingly, we attempt to train GANs using a cooperative dual evolutionary paradigm, which is effective for large-scale optimization task and diversity learning \cite{Goh2009ACC,Omidvar2014CooperativeCW,He2016CooperativeCM,Lu2018CooperativeCD,Zhang2019DynamicCC,Zou2019ADM,Ding2019DeepNC,Zhang2020NovelPS}.
	
	In light of the above observation, we propose a cooperative dual evolution based generative adversarial network, called \textit{CDE-GAN}, to train the model steadily and improve generation performance effectively. In essence, CDE-GAN  incorporates dual evolution with respect to the generator(s) and discriminators into a unified evolutionary adversarial framework to conduct effective adversarial multi-objective optimization. Thus, it exploits the complementary properties and injects dual mutation diversity into training to steadily diversify the estimated density in capturing multi-modes and to improve the generative performance. Our strategy towards achieving this goal is to decompose the complex adversarial optimization problem into two subproblems (generation and discrimination), and each subproblem is solved with a separated subpopulation (\textit{E-Generators} and \textit{E-Discriminators}), evolved by its own evolutionary algorithm, including individual variations (mutations), evaluation (fitness function), and selection. Unlike the existing EC based GANs \cite{Schmiedlechner2018TowardsDC,Wang2019EvolutionaryGA,Toutouh2019SpatialEG}, CDE-GAN simultaneously evolves a population of generators and an array of discriminators by operating their various objective functions (\textit{mutations}) that are interpretable and complementary. During training E-Generators, acting as parents, generators of CDE-GAN undergo different mutations to produce offspring to adapt to the dynamic environment (E-Discriminators). Meanwhile, we term the quality and diversity of samples generated by the evolved offspring as a fitness score for evaluating the offspring's performance. According to the fitness score, poorly performing offspring are removed, and the remaining well-performing offspring are preserved and used for next-generation training (i.e., evolution). Given optimal generators, the similar mechanism holds for E-Discriminators with its own evolutionary algorithm, and thus discriminators provide more informative gradient to generators for distribution diversity learning\footnote{Distribution diversity learning denotes that generator of GAN can learn the diversity distribution that covers different data modes of true data distribution \cite{Goodfellow2016NIPS2T,Tolstikhin2017AdaGANBG}.}. To keep the balance between E-Generators and E-Discriminators, we proposed a \textit{Soft Mechanism} to bridge them to conduct effective and stable adversarial training. In this way, CDE-GAN possesses three key benefits: 1)  the cooperative dual evolution (E-Generators and E-Discriminators) injects diversity into training so that CDE-GAN can cover different data modes, which significantly mitigates mode collapse of GANs; 2) CDE-GAN terms adversarial training as an adversarial multi-objective optimization problem, the multiple discriminators provide informative feedback gradient to generators for stabilizing the training process; 3) the complementary mutations in E-Generators and E-Discriminators will help model place fair distribution of probability mass across the modes of the data generating distribution.

	To summarize, this study makes the following salient contributions:
	
	\begin{itemize}
			\item  We propose a novel GAN method, termed cooperative dual evolution based generative adversarial network (CDE-GAN), to circumvent adversarial optimization difficulties of GANs, i.e., mode collapse and instability. To achieve this goal, CDE-GAN  incorporates dual evolution with respect to the generator(s) and discriminators into a unified evolutionary adversarial framework to conduct effective adversarial multi-objective optimization. Thus it exploits the complementary properties and injects dual mutation diversity into training to steadily diversify the estimated density in capturing multi-modes and improve generative performance.

			\item We design \textit{E-Generators} and \textit{E-Discriminators}, which evolve respectively by their own evolutionary algorithms, to solve the two subproblems (i.e., generation and discrimination) of CDE-GAN. To keep the balance between E-Generators and E-Discriminators, we further propose a \textit{Soft Mechanism} to cooperate them to conduct effective adversarial training.

		\item  We carry out extensive experiments on several benchmark datasets to demonstrate that our proposed method achieved obvious advantages over the existing methods, which proves the
		superiority and great potentials of CDE-GAN.
	\end{itemize}

	The remainder of this paper is organized as follows. Section \ref{sec2} gives related works in the field of generative adversarial networks. The proposed CDE-GAN is illustrated in Section \ref{sec3}. The performance and evaluation are provided in Section \ref{sec4}. Section \ref{sec5} presents the discussion. Section \ref{sec6} provides a summary and the outlook for future research.

	\section{Related Work}\label{sec2}

	In the following, we provide a review of GANs that developed various adversarial training approaches to overcome the optimization difficulties based on different methods.
	
	\subsection{Modifying Training Objective for GANs}\label{sec2.1}
	Modifying training objective of GANs is a typical way to improve and stabilize the optimization of GANs. Radford \textit{et al.} \cite{Radford2015UnsupervisedRL} introduced DCGAN to improve training stability. In \cite{Arjovsky2017WassersteinGA}, Arjovsky proposed Wasserstein-GAN to minimize a reasonable and efficient approximation of the Earth Mover (EM) distance for promoting the stability of training, and  showed the corresponding optimization problem theoretically. Meanwhile, Metz \textit{et al.} \cite{Metz2017UnrolledGA} introduced a method to stabilize GANs and increase diversity by defining the generator objective with respect to an unrolled optimization of the discriminator. Based on the idea that gradient signals from Denoising AutoEncoder (DAE) can guide the generator towards producing samples whose activations are close to the manifold of real data activations, Denoising Feature Matching (DFM) is proposed to improve GANs training \cite{WardeFarley2017ImprovingGA}. SN-GAN called spectral normalization to stabilize the discriminator's training and achieved promising generative performance \cite{Miyato2018SpectralNF}. Although some of these methods are practically and theoretically well-founded, convergence remains elusive in practice.
	
	\subsection{Multi-discriminator for GANs}\label{sec2.2}
	Unlike the traditional GANs performed single-objective optimization, some works attempt to revisit the multiple discriminators setting by framing the simultaneous optimization of different discriminator models as a multi-objective optimization problem. Thus it will overcome the problem of lacking informative gradient signal provided by the discriminators. Nguyen \textit{et al.} \cite{Nguyen2017DualDG} proposed D2GAN, combing the KL and reverse KL divergences into a unified objective function, to exploits the complementary statistical properties from these divergences to diversify the estimated density in capturing multi-modes effectively. Durugkar \textit{et al.} \cite{Durugkar2017GenerativeMN} simultaneously introduced multiple discriminators into the GAN framework to weaken the discriminators of GMAN, which provides informative feedback to generator and better guides generator towards amassing distribution in approximately true data region. In \cite{Neyshabur2017StabilizingGT}, Neyshabur proposed an array of discriminators, each of which looks at a different random low-dimensional projection of the data, to play the adversarial game with a single generator. Thus the individual discriminators are failed to reject generated samples perfectly. In \cite{Doan2019OnlineAC}, Doan argued that less expressive discriminators are smoother and have a general coarse-grained view of the mode's map, which enforces the generator to cover a wide region of the true data distribution.  Albuquerque \textit{et al.} \cite{Albuquerque2019MultiobjectiveTO} framed the training of multi-discriminator based GANs as a multi-objective optimization problem and analyzed its effectiveness. Although multi-discriminator based GANs perform promising results, they neglect to mine the prominence of the generator further.

	\subsection{Multi-generator for GANs}\label{sec2.3}
	Multi-generator based GANs simultaneously trained multiple generators with the object that a mixture of their induced distributions would approximate the data distribution. Thus, the mode collapse problem can be alleviated. Motivated by the boosting method, Tolstikhin \textit{et al.} \cite{Tolstikhin2017AdaGANBG} trained a mixture of generators by sequentially training and adding new generators to the mixture. Arora \textit{et al.} \cite{Arora2017GeneralizationAE} introduced a MIX+GAN framework to optimize the minimax game with the reward function being the weighted average reward function between any pair of generator and discriminator. In \cite{Ghosh2018MultiagentDG}, Ghost proposed MAD-GAN that trains a set of generators using a multi-class discriminator, which predicts which generator produces the sample and detecting whether the sample is fake. Additionally, MGAN \cite{Hoang2018MGANTG} was developed to overcome the mode collapsing problem, and its theoretical analysis was provided. Indeed, multi-generator based GANs break the balance of generator and discriminator. Thus, the additional supervised information is typically used to steady adversarial training of GANs.

	\subsection{Evolutionary Computation for GANs}\label{sec2.4}
	In fact, the aforementioned GAN methods are limited by the specified adversarial optimization strategy, which heavily limits optimization performance during training. Since EC has been successfully applied to solve many deep learning challenges \cite{Sun2019EvolvingUD,Sun2019APS,Sun2020EvolvingDC,Sun2020SurrogateAssistedED,Toutouh2020RepurposingHG}, some researchers attempt to overcome different training problems of GANs using the EC technique. In \cite{Schmiedlechner2018TowardsDC}, Schmiedlechner proposed Lipizzaner, which provides population diversity by training a two-dimensional grid (each cell contains a pair of generator-discriminator) of GANs with a distributed evolutionary algorithm. Wang \textit{et al.} \cite{Wang2019EvolutionaryGA} introduced E-GAN to inject mutation diversity into adversarial optimization of GANs by training the generator with three independent objective functions then selecting the resulting best performing generator for the next batch. Based on E-GAN, Liu \textit{et al.} developed CatGAN with hierarchical evolutionary learning for category text generation. In \cite{Toutouh2019SpatialEG}, Toutouh proposed Mustangs, hybridizing E-GAN and Lipizzaner, to combine mutation and population approaches to diversity improvement of GANs. Costa \textit{et al.} \cite{Costa2019COEGANET} developed COEGAN, using neuro-evolution and coevolution in the GAN training, to provide a more stable training method and the automatic design of neural network architectures. However, the aforementioned evolutionary GANs evolve a population of generators or GANs to play the adversarial game, which will result that a GAN evolves in a static environment. Thus the evolutionary dynamic of GANs is limited, and the balance between generator and discriminator of GANs is unstable. To this end, we propose a cooperative dual evolution based GAN to conduct effective adversarial multi-objective optimization. It exploits the complementary properties and injects dual mutation diversity into training to steadily diversify the estimated density in capturing multi-modes and to improve the generative performance.
	
	\section{Proposed Method}\label{sec3}
	
	\begin{figure*}[t]
		\centering
		\includegraphics[width=18.0cm,height=5.7cm]{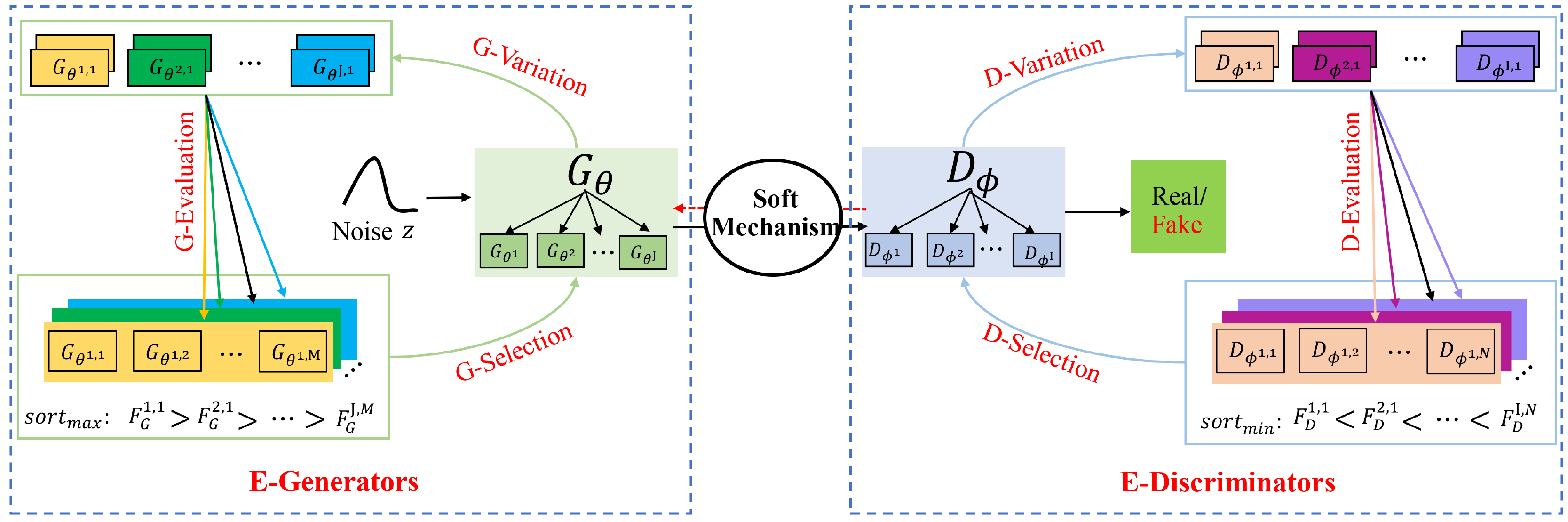}
		\caption{The pipeline of CDE-GAN. In brief, CDE-GAN decomposes the complex adversarial optimization problem into two subproblems (generation and discrimination), and each subproblem is solved with a separated subpopulation (i.e., \textit{E-Generators} and \textit{E-Discriminators}), evolved by its own evolutionary algorithm. The best offspring of E-Generators and E-Discriminator are served as new parents to produce the next generation's offspring during training. Furthermore, a \textit{Soft Mechanism} is proposed to cooperate E-Generators and E-Discriminators to conduct effective adversarial training.}
		\label{fig:pipeline}
	\end{figure*}
	
	Motivated by the success of cooperative co-evolutionary algorithm in large-scale optimization and diversity learning \cite{Goh2009ACC,Omidvar2014CooperativeCW,He2016CooperativeCM,Lu2018CooperativeCD,Zhang2019DynamicCC,Zou2019ADM,Ding2019DeepNC,Zhang2020NovelPS}, in this paper, we propose a \textit{Cooperative Dual Evolution based Generative Adversarial Network} (CDE-GAN) to circumvent drawbacks (i.e., instability and mode collapse) of GANs. In essence, CDE-GAN  incorporates dual evolution with respect to generator and discriminator into a unified evolutionary adversarial framework to conduct effective adversarial multi-objective optimization. Thus it exploits the complementary properties and injects dual mutation diversity into training to steadily diversify the estimated density in capturing multi-modes and improve the generative performance. As shown in Figure \ref{fig:pipeline}, CDE-GAN decomposes the complex adversarial optimization problem into two subproblems (generation and discrimination), and each subproblem is solved with a separated subpopulation (i.e.,  \textit{E-Generators} and \textit{E-Discriminators}), evolved by its own evolutionary algorithm during training. Specifically, we first propose a \textit{Soft Mechanism} to keep the balance between E-Generators and E-Discriminators and cooperate them to conduct effective adversarial training. Then, we introduce E-Generators and E-Discriminators, including their own \textit{Variations}, \textit{Evaluation} and \textit{Selection}.
	
	\subsection{Revisiting GANs}
	A GAN \cite{Goodfellow2014GenerativeAN} consists of a framework describing the interaction between two different models, i.e., generator ($G$) and discriminator ($D$), which are used to solve a min-max game optimization problem. Taking noisy sample $z \sim p_z$ as input, $G$ tries to learn real data distribution $p_{data}(x)$ by generating realistic looking samples $x \sim p_G(x)$ that are able to fool $D$ with $\min\limits_{G}$, while $D$ attempts to differentiate between samples from the data distribution and the ones produced by $G$ with $\max\limits_{D}$. Mathematically, GAN \cite{Goodfellow2014GenerativeAN} is formulated as 
		\begin{gather}
		\label{eq:gan}
		\min _{G} \max _{D} \mathbb{E}_{x \sim p_{\text {data }}}[\log D(x)]+\mathbb{E}_{z \sim p_{z}}[\log (1-D(G(z)))],
		\end{gather} 
	where $D(x)$ is the output of $D$ when the input is real data, $D(G(z))$ is the output of $D$ when the input is the data generated by $G$.  Most existing GANs conduct a similar adversarial procedure with different adversarial objectives, the optimization objective of CDE-GAN is presented in Section \ref{sec3.4}.
	
	\subsection{Soft Mechanism}\label{sec3.1}
	For the sake of keeping the balance between E-Generators and E-Discriminators, we proposed a \textit{Soft Mechanism} to bridge them to conduct effective adversarial training. In practice, the generator's learning will be impeded when it trains over a far superior discriminator. That is, the generator is unlikely to generate any samples considered "realistic" according to the discriminator's standards, and thus the generator will receive uniformly negative feedback \cite{Durugkar2017GenerativeMN}. This is problematic for GANs because the information contained in the gradient derived from negative feedback only dictates where to drive down the generated distribution $p_G(x)$, not specifically where to increase $p_G(x)$. Furthermore, it inevitably increases $p_G(x)$ in other regions of true data distribution $p_{data}(x)$ to maintain $\int_{\mathcal{X}} p_{G}(x)=1$, which may or may not contain samples from the true dataset (\textit{whack-a-mole} dilemma). To this end, the generator and discriminator can not be well balanced for effective adversarial training. In fact, the degenerate results of GANs can be avoided by employing learner (discriminator) with limited capacity and corrupting data samples with noise \cite{Neyshabur2017StabilizingGT,Durugkar2017GenerativeMN,Doan2019OnlineAC}. To this end, a generator is more likely to meet positive feedback against a more lenient discriminator, which may better guide a generator towards amassing $p_G(x)$ in approximately correct regions of $p_{data}(x)$.
		
	Inspired by this observation, we use a \textit{Soft Mechanism} of the classical Pythagorean method parameterized by $\delta$ to \textit{soften} the maximization of discriminators. Thus discriminators of CDE-GAN will be well weaken, which avoids the problem of whack-a-mole dilemma during training and enables us to obtain a diverse set of seemingly tenable hypotheses for the true data distribution $p_{data}(x)$. Furthermore, using \textit{softmax} has a well-known benefit of being differentiable. Specifically, given the optimal E-Discriminators $D_\phi^*$, the generator(s) {$G_{\theta^j}$} trains against a \textit{softmax} weighted arithmetic average of $I$ different discriminators. It can be formulated as
	\begin{gather}
	\label{eq:soft}
	\mathcal{L}_{G_{\theta^j}}(\delta)=\sum_{i=1}^{I} w_{i} l_i,
	\end{gather} 
	where $w_{i}=\frac{e^{\delta l_i}}{\sum_{t=1}^{I} e^{\delta l_t}}$, $l_i= \mathcal{M}_{G}(D_{\phi^i})$, $\mathcal{M}_{G}\in\{\mathcal{M}_{G}^{minimax},\mathcal{M}_{G}^{heuristic},\mathcal{M}_{G}^{ls}\}$ is the variation (mutation) of the evolutionary algorithm in E-Generators (see Section \ref{sec3.3.1} for more details), and $\delta \ge 0$. When $\delta \rightarrow \infty$, $G_{\theta^j}$ trains against only a single weak discriminator; when $\delta = 0$, $G_{\theta^j}$ trains against an ensemble with equal weights. These two scenes are not desired. Here, we set $\delta=1$ for our all experiments.

	\subsection{E-Generators}\label{sec3.2}
	In fact, the most existing GAN methods (e.g., modifying training objective based GANs, multi-discriminator based GANs, and multi-generator based GANs) are limited by the specified adversarial optimization strategy, which heavily affects optimization performance during training. To this end, we build \textit{E-Generators} that evolves a set of generators (parents) $\{G_{\theta^{1}},G_{\theta^{2}},\cdots,G_{\theta^{J}}\}$ with number of $J$ in a given dynamic environment (\textit{E-Discriminators}) based on an evolutionary algorithm to produce a set of new generators (offspring) $\{G_{\theta^{1,1}},G_{\theta^{1,2}},\cdots,G_{\theta^{1,M}}, \cdots, G_{\theta^{J,1}},G_{\theta^{J,2}},\cdots,G_{\theta^{J,M}}\}$. E-Generators is an individual subpopulation to solved a subproblem (generation), which is cooperative with E-Discriminators to solve the adversarial multi-objective optimization problem of GANs. The optimization of E-Generators is formulated as
		
		\begin{gather}
		\label{eq:G_objective}
		\min\mathcal{L}_{G_\theta}=\left[\mathcal{L}_{G_{\theta^1}}, \mathcal{L}_{G_{\theta^2}}, \cdots, \mathcal{L}_{G_{\theta^J}}\right]^{T},
		\end{gather} 
		where $\mathcal{L}_{G_{\theta^j}}$ is defined in Eq. \ref{eq:soft}, and $\theta =\{\theta ^1,\theta ^2,\cdots ,\theta ^J\}$.
		
		Based on the evolutionary algorithms, we take \textit{G-Variations} (mutations), \textit{G-Evaluation} (fitness function) and \textit{G-Selection} (selection) for evolving E-Generators. The evolutionary process of E-Generators is presented in Fig. \ref{fig:pipeline} and Algorithm \ref{algotithm:CDE-GAN}.
	
	\begin{algorithm*}[t]
		
		\caption{The algotithm of CDE-GAN.}
		\label{algotithm:CDE-GAN}
		\begin{algorithmic}[1]
			\Require  The generator $G_\theta$; the discriminator $D_\phi$; the number of iterations $T$; the discriminator's updating steps per iteration $K$; the number of parents for E-Generator $J$; the number of parents for E-Discriminator $I$; the number of mutations for generator $M$; the number of mutations for discriminator $N$; the hyper-parameter $\gamma$ of fitness function of E-Generators.*
				\State Initialize generators' parameter $\{\theta^{1}, \theta^{2}, \cdots, \theta^{J}\}$, initialize discriminators' parameter $\{\phi^{1}, \phi^{2}, \cdots, \phi^{I}\}$.
				\For{$t=1,\cdots,T$}
				\hspace{1cm}\For{$k=1,\cdots,K$} 	
				\For{$i=1,\cdots,I$}  \Comment{ E-Discriminators Evolution}
				\State Sample a batch of $x_{real}\sim p_{data}$.
				\State Sample a batch of $z \sim p_z$, and generate a batch of $x_{fake}$ with E-Generators.
				\State $D_{\phi^i}$ produces $N$ offspring $D_{\phi^{i,n}}$ via D-Variation, that is, updating $D_{\phi^i}$ via Eq. \ref{eq:D_max} and  Eq. \ref{eq:D_ls}  respectively. \Comment{D-Variation}
				\State Evaluate the $N$ evolved offspring of E-Discriminators  via Eq. \ref{eq:f_D}. \Comment{ D-Evaluation}
				\EndFor
				\State Select the best-performing offspring $\{D_{\phi^{1,1}}, D_{\phi^{2,1}},\ldots, D_{\phi^{I,1}}\}$ as next generation's parents of E-Discriminators via Eq. \ref{eq:d-sele} and Eq. \ref{eq:d-sort}. \Comment{ D-Selection}
				\EndFor
				\For{$j=1,\cdots,J$} \Comment{E-Generators Evolution}
				\State Sample a batch of $z \sim p_z$.
				\State $G_{\theta^j}$ produces $M$ offspring $G_{\theta^{j,m}}$ via G-Variation, that is, updating $G_{\theta^j}$ via Eq. \ref{eq:g_minimax}, Eq. \ref{eq:heuristic} and  Eq. \ref{eq:g_ls}  respectively. \Comment{G-Variation}
				\State Evaluate the $M$ evolved offspring of E-Generators via Eq. \ref{eq:f}. \Comment{ G-Evaluation}
				\EndFor
				\State Select the best-performing offspring $\{G_{\theta^{1,1}}, G_{\theta^{2,1}},\ldots, G_{\theta^{J,1}}\}$ as next generation's parents of E-Generators via Eq. \ref{eq:g-sele} and Eq. \ref{eq:g-sort}. \Comment{ G-Selection}
				\EndFor
		\end{algorithmic}
		*Default values: $B=32$, $K=3$, $M=3$ and $N=2$.
	\end{algorithm*}

\subsubsection{G-Variation}\label{sec3.3.1}
	CDE-GAN applies three complementary mutations corresponding with three different minimization objective w.r.t generator for the evolution of E-Generators, i.e., \textit{G-Minimax mutation} ($\mathcal{M}_{G}^{minimax}$), \textit{G-Heuristic mutation} ($\mathcal{M}_{G}^{heuristic}$), and \textit{G-Least-Square mutation} ($\mathcal{M}_{G}^{ls}$). They are corresponding to vanilla GAN (GAN) \cite{Goodfellow2014GenerativeAN}, non-saturated GAN (NS-GAN) \cite{Goodfellow2014GenerativeAN}, and least square GAN (LSGAN) \cite{Mao2017LeastSG,Mao2019OnTE}, respectively. In contrast to mutations of E-GAN involving generator(s) against a single specified discriminator, the mutations of E-Generators train over multiple evolutionary discriminators. The G-Minimax mutation corresponds to the minimization objective of the generator in the vanilla GAN \cite{Goodfellow2014GenerativeAN}. According to Eq. \ref{eq:soft}, it is defined as
	\begin{gather}
	\label{eq:g_minimax}
	\mathcal{M}_{G}^{minimax}=\frac{1}{2} \sum_{i=1}^{I} w_{i} \mathbb{E}_{z \sim p_{z}}[\log (1-D_{\phi^i}(G_\theta(z)))].
	\end{gather}
	In fact, G-Minimax mutation is to minimize the Jensen-Shannon Divergence (JSD) between the data and model distributions. If discriminators distinguish generated samples with high confidence, the gradient tend to vanish, G-Minimax mutation fails to work; if discriminators cannot completely distinguish real/fake sample, the G-Minimax mutation will provide informative gradient for adversarial training. Thus, G-Minimax mutation typically evolves the best offspring in the latter training process for CDE-GAN. 
	
	Additionally, G-Heuristic mutation is non-saturating when the discriminator convincingly rejects the generated samples, and thus it avoids gradient vanish. According to Eq. \ref{eq:soft}, it is formulated as follow: 
	\begin{gather}
	\label{eq:heuristic}
	\mathcal{M}_{G}^{heuristic}=-\frac{1}{2} \sum_{i=1}^{I} w_{i} \mathbb{E}_{z \sim p_{z}}[ \log(D_{\phi^i}(G_\theta(z)))].
	\end{gather}
	However, G-Heuristic mutation may direct to training instability and generative quality fluctuations because it pushes the data and model distributions away each other. 
	
	As for G-Least-Square mutation, it is inspired by LSGAN \cite{Mao2017LeastSG}, which applies this criterion to adapt both generator and discriminator. According to Eq. \ref{eq:soft}, it can be written as
	\begin{gather}
	\label{eq:g_ls}
	\mathcal{M}_{G}^{ls}=\sum_{i=1}^{I} w_{i} \mathbb{E}_{z \sim p_{z}}[ (D_{\phi^i}(G_\theta(z)-1)^2].
	\end{gather}
	Similar to G-Heuristic, G-Least-Square mutation will effectively avoid gradient vanish when the discriminator easily recognizes the generated samples. Meanwhile, G-Least-Squares mutation will partly avoid mode collapse, because it neither assigns an extremely high cost to generate fake samples nor assigns an extremely low cost to mode dropping. Thus, these different mutations provide various training strategies for E-Generators, which injects mutation diversity into training to diversify the estimated density in capturing multi-modes and constructs a complementary population of generators for steady training. See \cite{Goodfellow2014GenerativeAN,Mao2017LeastSG,Arjovsky2017TowardsPM,Wang2019EvolutionaryGA} for more corresponding theoretical analysis of these initial objective functions.

	\subsubsection{G-Evaluation} 
		After producing the offspring with different mutations, we evaluate the individual's quality for each child using a \textit{fitness} function $\mathcal{F}_G$ that depends on the current environment (i.e., E-Discriminators $D_\phi$). Considering two typical properties (quality\footnote{the generated samples are so realistic enough that it will fool the superior discriminator.\cite{Salimans2016ImprovedTF}} and diversity\footnote{the model distribution is more possible to cover the real data distribution. It could largely avoid mode collapse.\cite{Heusel2017GANsTB}}) of generated sample, $\mathcal{F}_G$ consists of two fitness scores, i.e., \textit{quality fitness scores} $\mathcal{F}_{G^q}$ and \textit{diversity fitness score} $\mathcal{F}_{G^d}$, for evaluating the performance of the offspring (generators) of E-Generators. On the one hand, the generated samples produced by generator are feed into discriminators and the sum value of the output are calculated, which is termed as $\mathcal{F}_{G^q}$:
	\begin{gather}
	\label{eq:f_q}
	\mathcal{F}_{G^q}=\sum_{i=1}^{I} w_{i}\mathbb{E}_{z \sim p_{z}}[D_{\phi^i}(G_\theta(z))].
	\end{gather}
	The higher quality score generators achieved, the more reality generated samples gotten. Reflecting the quality performance of generators at each evolutionary step, discriminators are constantly upgraded to be optimal during the training process. On the other hand, we also focus on the
	diversity of generated samples and attempt to gather a better group of generators to circumvent the mode collapse issue in adversarial optimization of GANs. According to \cite{Nagarajan2017GradientDG}, a gradient-based regularization term can stabilize the GAN optimization and suppress mode collapse. To this end, the minus log-gradient-norm of optimizing $D_\phi$ is used to measure the diversity fitness score of generated samples
	\begin{gather}
	\label{eq:f_d}
	\begin{aligned}
	\mathcal{F}_{G^d}&=-\sum_{i=1}^{I}w_{i} (\log ||\nabla_{D_{\phi^i}}-\mathbb{E}_{x \sim p_{data}}[\log D_{\phi^i}(x)]\\
	&-\mathbb{E}_{z \sim p_{z}}[\log (1-D_{\phi^i}(G_\theta(z)))]||).\end{aligned}
	\end{gather}
	
	When an evolved generator obtains a relatively max value, which corresponds to small gradients of $D_\phi$, its generated samples
	tend to spread out enough and to avoid the discriminators from having obvious countermeasures. Therefore, $\mathcal{F}_G$ is formulated as
	\begin{gather}
	\label{eq:f}
	\mathcal{F}_G=\mathcal{F}_{G^q}+\gamma\mathcal{F}_{G^d},
	\end{gather}
	$\gamma \geq 0$ is used for balancing the quality and diversity of generated samples. To this end, the performance $F_G^{j,m}$ of each evolved offspring $G_{\theta^{j,m}}$ is evaluated using $\mathcal{F}_G$. Generally, a max fitness score
	$\mathcal{F}_G$ directs to higher training efficiency and better generative performance.
	
	\subsubsection{G-Selection} 
		
		Finally, a simple yet useful survivor selection strategy $(\mu,\lambda)$-selection \cite{Kramer2016Machine} is employed to select the new parents of next evolution according to the fitness score $\mathcal{F}_G$ of existing individuals. The selection function for the offspring of E-Generators is defined  as
		\begin{gather}
		\label{eq:g-sele}
		\left\{\mathcal{F}_G^{1, 1}, \mathcal{F}_G^{2, 1}, \ldots, \mathcal{F}_G^{J, M}\right\} \leftarrow sort_{max}\left(\left\{\mathcal{F}_G^{j, m}\right\}\right).
		\end{gather}
		After sorting, $J$ individuals $\{G_{\theta^{1}},G_{\theta^{2}},\cdots,G_{\theta^{J}}\}$ possessing the \textit{maximum} fitness score can be survived for next evolution during adversarial training. It is formulated as 
		\begin{gather}
		\label{eq:g-sort}
		\theta^{1}, \theta^{2}, \ldots, \theta^{J} \leftarrow \theta^{1, 1}, \theta^{2, 1}, \ldots, \theta^{J, 1}.
		\end{gather}

	\subsection{E-Discriminators}\label{sec3.3}
	Recently, some works attempt to revisit the multiple discriminators setting by framing the simultaneous optimization of different discriminator models as a multi-objective optimization problem \cite{Nguyen2017DualDG,Durugkar2017GenerativeMN,Neyshabur2017StabilizingGT,Doan2019OnlineAC}. Thus it would overcome the problem of lacking informative gradient signal provided by discriminators. To this end, we develop E-Discriminators, which evolves a population of discriminators (parents) $\{D_{\phi^{1}},D_{\phi^{2}},\cdots,D_{\phi^{I}}\}$ to a set of new discriminators (offspring) $\{D_{\phi^{1,1}},D_{\phi^{1,2}},\cdots,D_{\phi^{1,N}}, \cdots, D_{\phi^{I,1}},D_{\phi^{I,2}},\cdots,D_{\phi^{I,N}}\}$ using its individual evolutionary algorithm. In fact, E-Discriminators possesses two advantages to help CDE-GAN steadily achieve promising generative performance: 1) the evolutionary mechanism provides a dynamic strategy to discriminators, thus the trade-off between generator(s) and discriminators is well adjusted during training; 2) individual discriminators are unable to reject generated samples perfectly and continue to provide informative gradients to the generator throughout training. Each discriminator $D_{\phi^i}$ of E-Discriminators maximizes its own objective function. Mathematically, we formulate E-Discriminators' objective as
		\begin{gather}
		\label{eq:D_objective}
		\max\mathcal{L}_{D_\phi}=\left[\mathcal{M}_{D_{\phi^1}}, \mathcal{M}_{D_{\phi^2}}, \cdots, \mathcal{M}_{D_{\phi^I}}\right]^{T},
		\end{gather} 
		where $ \mathcal{M}_{D_{\phi^i}}\in\{\mathcal{M}_{D}^{minimax},\mathcal{M}_{D}^{ls}\}$ is the variation (mutation) of the evolutionary algorithm in E-Discriminators, and $\phi =\{\phi ^1,\phi ^2,\cdots ,\phi ^I\}$. 
		
		Specifically, given the optimal ${G_\theta^*}$, E-Discriminators evolves with its own \textit{D-Variation}, \textit{D-Evaluation} and \textit{D-Selection} during training. The evolutionary process of E-Discriminators is presented in Fig. \ref{fig:pipeline} and Algorithm \ref{algotithm:CDE-GAN}.
	
	\subsubsection{D-Variation}
		Here, we take various objective functions as the mutations for E-Discriminators. To keep the objective functions of E-Discriminators are corresponding to the objective functions of E-Generators, we take two different objective w.r.t  discriminators, including \textit{D-Minimax mutation} ($\mathcal{M}_{D}^{minimax}$), and \textit{D-Least-Square mutation} ($\mathcal{M}_{D}^{ls}$). According to \cite{Goodfellow2014GenerativeAN}, D-Minimax mutation is a shared objective function for vanilla GAN and NS-GAN, it can be formulated as
	\begin{gather}
	\label{eq:D_max}
	\begin{aligned}
	\mathcal{M}_{D}^{minimax}=&\mathbb{E}_{x \sim p_{ data }}[\log D_{\phi^i}(x)]\\
	&+\mathbb{E}_{z \sim p_{z}}[\log (1-D_{\phi^i}(G_\theta(z)))].
	\end{aligned}
	\end{gather}
	Indeed, D-Minimax mutation adopts the sigmoid cross-entropy loss for the discriminator, which will lead to the problem of vanishing gradients when updating the generator using the fake samples that are on the correct side of the decision boundary, but are still far from the real data \cite{Mao2019OnTE}. Interestingly, the non-saturating loss $\mathcal{M}_{G}^{heuristic}$ will saturate when the input is relatively large; the minimax objective $\mathcal{M}_{G}^{minimax}$ will saturate when the input is relatively small. Thus, the generative performance of E-GAN \cite{Wang2019EvolutionaryGA} will be limited when it bases on single discriminator with minimax objective. In light of this observation, we further adopt least squares objective function for the second mutation of E-Discriminators:
	\begin{gather}
	\label{eq:D_ls}
	\begin{aligned}
	\mathcal{M}_{D}^{LS} &=\frac{1}{2} \mathbb{E}_{x} \sim p_{data}(x)\left[(D_{\phi^i}(x)-1)^{2}\right] \\
	&+\frac{1}{2} \mathbb{E}_{\boldsymbol{z} \sim p_{z}(z)}\left[(D_{\phi^i}(G_\theta(z)))^{2}\right]
	\end{aligned}
	\end{gather}
	Benefiting the least square objective function penalizes samples that lie in a long way on the correct side of the decision boundary, $\mathcal{M}_{D}^{LS}$ is capable of moving the fake samples toward the decision boundary. To this end, $\mathcal{M}_{D}^{LS}$ will help CDE-GAN generate samples that are closer to the real data. Overall, $\mathcal{M}_{D}^{minimax}$ and $\mathcal{M}_{D}^{LS}$ are complementary to each other and will provide a promising optimization direction for the evolution of E-Discriminators, which effectively adjusts the trade-off between generator(s) and discriminators.

	\subsubsection{D-Evaluation}
		To evaluate the subpopulation evolved by E-Discriminators, we take the minus log-gradient-norm of optimizing each discriminator $D_{\phi^i}$ as fitness function $\mathcal{F}_{D}$, which is corresponding to the diversity fitness score of generated samples in E-Generators:
	\begin{gather}
	\label{eq:f_D}
	\begin{aligned}
	\mathcal{F}_{D}&=-\log ||\nabla_{D_{\phi^i}}-\mathbb{E}_{x \sim p_{data}}[\log D_{\phi^i}(x)]\\
	&-\mathbb{E}_{z \sim p_{z}}[\log (1-D_{\phi^i}(G_\theta(z)))]||\end{aligned}
	\end{gather}
	There are two reasons for this setting: 1) the gradient reveals the train status of GANs. When the generator can generate realistic samples, the discriminators will not reject the generated sample confidently (i.e., $D_{\phi^i}$ updated with small gradient); when the generator collapses to a small region, the discriminator will subsequently label collapsed points as fake with obvious countermeasure (i.e., $D_{\phi^i}$ updated with big gradient); 2) the cooperative fitness function (Eq. \ref{eq:f_d} and Eq. \ref{eq:f_D}) between E-Generators and E-Discriminators will keep the adversarial consistency of them, which will improve the training stability of CDE-GAN. Therefore, $\mathcal{F}_{D}$ is effective for representing whether the model falls in mode collapse, and thus it guides E-Discriminators to evolve with the meaningful direction.

	\subsubsection{D-Selection}
		After evaluation, the new parents of next evolution in E-Discriminators can be selected following the principle of survival-of-the-fittest, which is similar to the selection of E-Generators. We define selection function for the offspring of E-Discriminators as
		\begin{gather}
		\label{eq:d-sele}
		\left\{\mathcal{F}_D^{1, 1}, \mathcal{F}_D^{2, 1}, \ldots, \mathcal{F}_D^{I, N}\right\} \leftarrow sort_{min}\left(\left\{\mathcal{F}_D^{i, n}\right\}\right).
		\end{gather}
		After sorting, $I$ individuals $\{D_{\phi^{1}},D_{\phi^{2}},\cdots,D_{\phi^{I}}\}$ possessing the \textit{minimum} fitness score can be survived for next evolution during adversarial training.  It can be written as 
		\begin{gather}
		\label{eq:d-sort}
		\phi^{1}, \phi^{2}, \ldots, \phi^{I} \leftarrow \phi^{1, 1}, \phi^{2, 1}, \ldots, \phi^{I, 1}.
		\end{gather}

	\subsection{Adversarial Multi-objective Optimization}\label{sec3.4}
	
	CDE-GAN is optimized during the iterative evolution (training) between E-Generators and E-Discrimiantors. Here, we further demonstrate the total optimization of CDE-GAN. In this work, E-Generators is trained with the multi-objective optimization (Eq. \ref{eq:G_objective}) when $J>1$; E-Discriminators is also trained with the multi-objective optimization (Eq. \ref{eq:D_objective})  when $I>1$ \cite{Albuquerque2019MultiobjectiveTO,Deb2001MultiobjectiveOU}. To this end, CDE-GAN terms the adversarial training as a adversarial multi-objective optimization, formulated as
	\begin{gather}
	\label{eq:CDE_objective}
	\begin{aligned}
	\min _{G_\theta} \max _{D_\phi}&=\left[\mathcal{L}_{G_{\theta^1}}, \mathcal{L}_{G_{\theta^2}}, \cdots, \mathcal{L}_{G_{\theta^J}}\right]^{T}\\
	  				   		    	&+\left[\mathcal{M}_{D_{\phi^1}}, \mathcal{M}_{D_{\phi^2}}, \cdots, \mathcal{M}_{D_{\phi^I}}\right]^{T}.\end{aligned}
	\end{gather} 
	Each $\mathcal{L}_{G_{\theta^j}}$ and $\mathcal{M}_{D_{\phi^i}}$ are optimized individually during evolution. Since the E-Generators and E-Discriminators are evolved with their own evolutionary algorithms, the generator(s) and discriminators are dynamically optimized in each evolutionary step. Thus, CDE-GAN will well exploit the trade-off between E-Generators and E-Discriminators and to conduct stable and effective training. The multiple discriminators provide informative feedback gradient to generator(s) for stabilizing the training process.

	\begin{figure*}[t]
		\centering
		\includegraphics[width=6.5cm,height=5cm]{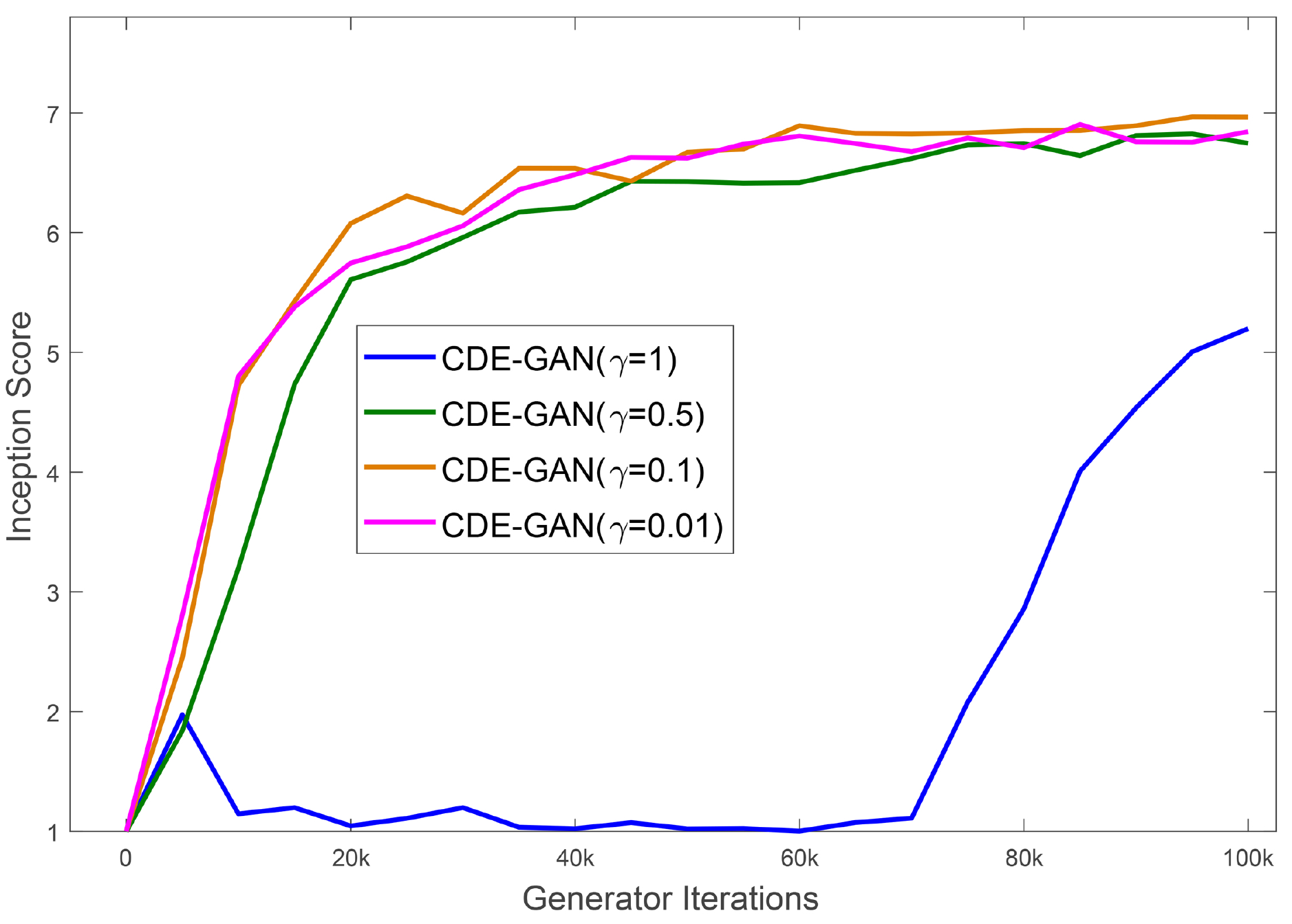}
		\hspace{1cm}
		\includegraphics[width=8cm,height=5cm]{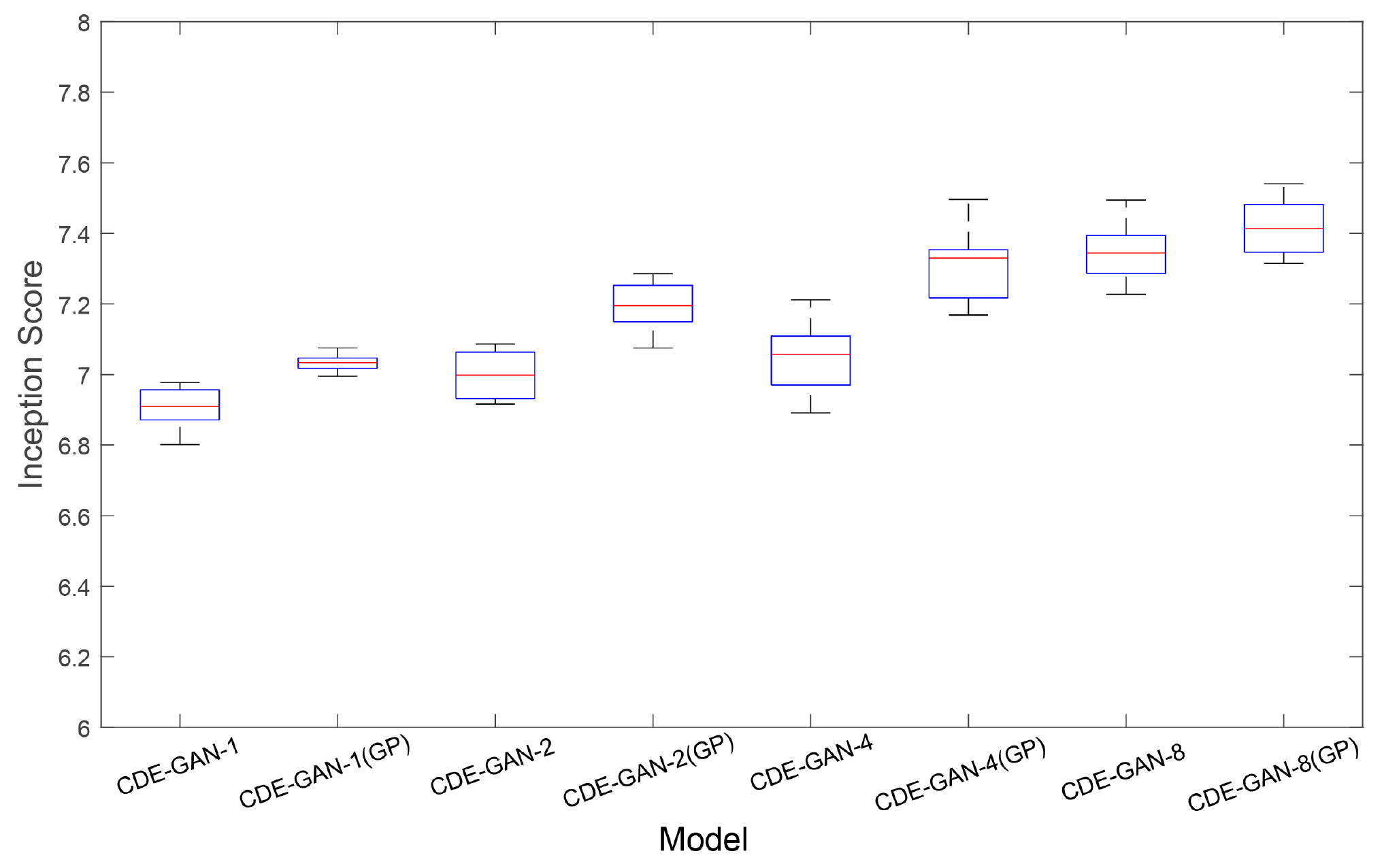}
		\\ \hspace{-0.1cm} (a) IS for various balance factor   \hspace{3.2cm} (b) IS for various numbers of discriminators \\ 
		\caption{Experiments on the CIFAR-10 dataset for hyper-parameters analysis. (a) Inception score evaluation for different CDE-GANs with various balance factor $\gamma=\{1,0.5,0.1,0.01\}$. (b) Inception score evaluation for different CDE-GANs with various numbers of discriminators $I=\{1,2,4,8\}$.}
		\label{fig:IS-dif-dis}
	\end{figure*}

	\begin{table*}[t]
		\centering
		\captionsetup{justification=centering}
		\caption{\textsc{Architectures of the Generative and Discriminative Networks Used in This Work, i.e., DCGAN Model, MLP with 3 Layers, and MLP with 4 Layers.} }\label{Table:network-architecture}
		\setlength{\tabcolsep}{8mm}{
			\begin{tabular}{l|l}
				\hline
				\textbf{Generative network}     &    \textbf{Discriminative network} \\
				\hline      
				\textbf{DCGAN}\cite{Radford2015UnsupervisedRL,Gulrajani2017ImprovedTO,Wang2019EvolutionaryGA}              &   \\ 
				\textbf{Input}: Noise $z \sim p_z$, 100                                     &  \textbf{Input}: Image, ($32 \times 32 \times 3$)               \\
				$[$layer 1$]$  Fully connected and Reshape to ($4\times4\times512$); ReLU;  &  $[$layer 1$]$ Convolution (4, 4, 128), stride=2; LeakyReLU; \\
				$[$layer 2$]$  Transposed Convolution (4, 4, 512), stride=2; ReLU;          &  $[$layer 2$]$ Convolution (4, 4, 256), stride=2; LeakyReLU; \\
				$[$layer 3$]$  Transposed Convolution (4, 4, 256), stride=2; ReLU;          &  $[$layer 3$]$ Convolution (4, 4, 512), stride=2; LeakyReLU; \\
				$[$layer 4$]$  Transposed Convolution (4, 4, 128), stride=2; ReLU;          &  $[$layer 4$]$ Fully connected (1); Sigmoid/Least squares;\\
				$[$layer 5$]$  Transposed Convolution (4, 4, 3), stride=2; Tanh;            &  \textbf{Output}: Real or Fake (Probability) \\
				\textbf{Output}: Generated Imgage, ($32 \times 32 \times 3$)                & \\
				\hline
				\textbf{MLP with 3 Layers} \cite{Mao2017LeastSG,Mao2019OnTE}                &    \\ 
				\textbf{Input}: Noise $z \sim p_z$, 256                                     &  \textbf{Input}: Point                           \\
				$[$layer 1$]$  Fully connected (128); ReLU;                                 &  $[$layer 1$]$ Fully connected (128); LeakyReLU; \\
				$[$layer 2$]$  Fully connected (128); ReLU;                                 &  $[$layer 2$]$ Fully connected (128); LeakyReLU; \\
				$[$layer 3$]$  Fully connected (2); Linear;                                 &  $[$layer 3$]$ Fully connected (1); Sigmoid/Least squares;       \\
				\textbf{Output}: Generated Point                                            &  \textbf{Output}: Real or Fake (Probability) \\
				\hline 
				\textbf{MLP with 4 Layers} \cite{Gulrajani2017ImprovedTO}                   &    \\ 
				\textbf{Input}: Noise $z \sim p_z$, 256                                     &  \textbf{Input}: Point                           \\
				$[$layer 1$]$  Fully connected (128); ReLU;                                 &  $[$layer 1$]$ Fully connected (128); ReLU; \\
				$[$layer 2$]$  Fully connected (128); ReLU;                                 &  $[$layer 2$]$ Fully connected (128); ReLU; \\
				$[$layer 3$]$  Fully connected (128); ReLU;                                 &  $[$layer 3$]$ Fully connected (128); ReLU; \\
				$[$layer 4$]$  Fully connected (2); Linear;                                 &  $[$layer 4$]$ Fully connected (1); Sigmoid/Least squares;  \\
				\textbf{Output}: Generated Point                                            &  \textbf{Output}: Real or Fake (Probability) \\
				\hline
				
		\end{tabular}}
	\end{table*}
	
	\section{Experiments and Evaluation}\label{sec4}
	In subsequent sections, we introduce the evaluation metrics, implementation details, and hyper-parameter analysis of experiments. Furthermore, we qualitatively and quantitatively analyze the generative performance of CDE-GAN to verify our claims. Finally, we demonstrate the advantages of our method over 11 state-of-the-art methods, including modifying training objective based GANs, multi-generator based GANs, multi-discriminator based GANs, and evolutionary computation based GANs.
	
	\subsection{Evaluation Metric}\label{sec4.0}
	We use the inception score (IS) \cite{Salimans2016ImprovedTF} to quantitatively evaluate the performance of the proposed method. It is a common quantitative evaluation metric in the image generation of GANs. IS uses the Inception model for every generated image to get the conditional label distribution $p(y|x)$. It is formulated as: $\exp \left(\mathbb{E}_{x} \operatorname{KL}(p(y|x) \| p(y))\right)$. IS takes simultaneously generative quality and diversity into account. The higher IS is achieved, the better quality and diversity of samples is generated. IS is calculated by Tensorflow code version using randomly generated 50k samples in this paper. Meanwhile, we also qualitatively evaluate our method with human visual conception.

	\subsection{Implementation Details}\label{sec4.1}
	In the following experiments, we use the default hyper-parameter values listed in Algorithm \ref{algotithm:CDE-GAN}. We conduct extensive experiments on one synthetic dataset (i.e., a mixture of 8 Gaussians arranged in a circle) and three real-word benchmark datasets (i.e., CIFAR10 \cite{Krizhevsky2016Learning}, LSUN-Bedrooms \cite{Yu2015LSUNCO}, and CelebA \cite{Liu2015DeepLF}) to prove the effectiveness of CDE-GAN. Furthermore, we adopt the same network architectures (DCGAN) with existing in works \cite{Radford2015UnsupervisedRL,Gulrajani2017ImprovedTO,Wang2019EvolutionaryGA} to conduct real data experiments for facilitating direct comparison. For the sake of fair comparison, we select MLP (with 3 layers \cite{Mao2017LeastSG,Mao2019OnTE}  or 4 layers \cite{Gulrajani2017ImprovedTO}) as the model architecture of CDE-GAN to conduct toy experiments and generate 512 points to cover modes. The model architectures are clearly displayed in Table \ref{Table:network-architecture}. The noisy vector $z$ is sampled from the uniform distribution $p_{z}$ with 100 dimensions and 256 dimensions for real-word datasets and synthetic datasets, respectively. We employ Adam optimizer with hyper-parameter ($\alpha=0.0002$, $\beta_1=0.5$, $\beta_2=0.99$) to optimize our model. All experiments are performed on a single NVIDIA 1080Ti graphic card with 11GB memory. We use PyTorch for the implementation of all the experiments.

	\subsection{Experiment 1: Hyper-Parameters Analysis}\label{sec4.2}
	In principle, there are two hyper-parameters closely corresponding to the performance of CDE-GAN, i.e., the balance factor $\gamma$ (see Eq. \ref{eq:f}) and the number of discriminators $I$. Since the hyper-parameter $\gamma$ is used for balancing the measurement of sample quality and diversity, it directs generator selection of E-Generators, and thus it will affect the effectiveness of CDE-GAN. Meanwhile, we analyze how the number of discriminators $I$ affects the sample diversity of the corresponding generator and select proper discriminators for balancing time consuming and generative performance of CDE-GAN.
	
	\subsubsection{Balance Factor $\gamma$}\label{sec4.2.1}
	In fact, the quality and diversity of the synthesized objects are two key goals of the generative task. Analogously, we also consider these two measurements for CDE-GAN evaluation in generation task. Here, we embed a balance factor $\gamma$ into the generator's fitness score to balance the quality and diversity of generated samples during generator updates. Similar to \cite{Wang2019EvolutionaryGA}, $\gamma>0$ is considered. If $\gamma$ is set as too small, the diversity fitness score is almost not considered; while $\gamma$ is set too large, the model is not stable since the gradient-norm of discriminators $D_\phi$ could vary largely. To this end, $0< \gamma\le 1$ is considered for the setting of our experiments.  
	
	To select a proper $\gamma$ for CDE-GAN, we run a grid search to find its value on CIFAR-10. As shown in Fig. \ref{fig:IS-dif-dis}(a), we take various balance factor $\gamma=\{1,0.5,0.1,0.01\}$ to conduct experiments. Results show that CDE-GAN is out of work at the beginning and achieves convergence with slow speed when $\gamma$ is set as 1; while it gains promising generative performance and comparable convergence speed, if $\gamma$ is set as a relatively small value, e.g., 0.01. Based on these observations, we take $\gamma=0.1$ to conduct later experiments on real-world datasets.

	\subsubsection{Number of Discriminators $I$}\label{sec4.2.2}
	Multi-discriminator based GANs frame GAN's training as a multi-objective optimization problem, which will overcome the problem of lacking informative gradient signal provided by the discriminator. Here, we can also extend CDE-GAN as a multi-discriminator based GAN. Specifically, we take E-Discriminators to evolve multiple discriminators ($I\geq 1$) with various mutations. According to \cite{Durugkar2017GenerativeMN,Albuquerque2019MultiobjectiveTO,Mordido2020microbatchGANSD}, the number of discriminators is closely corresponding to the diversity of the generated sample. Therefore, we take experiments to analyze how the number of discriminators affects the generative performance of CDE-GAN and select a promising $I$ for later experiments.  
	
	In Fig. \ref{fig:IS-dif-dis}(b), we report the box-plots of inception score evaluation for different CDE-GANs with different numbers of discriminators on CIFAR-10 across 3 independent runs. Results clearly show that increasing the number of discriminators yields better generative performance for CDE-GANs. Note that CDE-GAN is more stable when more numbers of discriminator are survived for evolution, because gradient penalty (GP) cannot largely improve the performance of CDE-GAN when more discriminators are set. The reason is that the generator will better fool the discriminators in the E-Discriminators when more numbers of discriminators are survived for evolution \cite{Neyshabur2017StabilizingGT,Albuquerque2019MultiobjectiveTO}. To this end, The E-Generators is more likely to meet positive feedback from E-Discriminators, which is benefiting for the stable training of CDE-GAN.. Moreover, we convince that CDE-GAN can further improve generative performance if more discriminators are survived during training. To take a trade-off between time consuming and efficacy, we conduct latter experiments with two discriminators for CDE-GAN, unless otherwise claimed.

	\subsection{Experiment 2: Generative Performance Evaluation}\label{sec4.3}
	In this section, we qualitatively and quantitatively analyze the generative performance of CDE-GAN to support our claims. Here, we take vanilla GAN (GAN) \cite{Goodfellow2014GenerativeAN}, non-saturated GAN (NS-GAN) \cite{Goodfellow2014GenerativeAN}, and least square GAN (LSGAN) \cite{Mao2017LeastSG,Mao2019OnTE}, wasserstein GAN (WGAN) \cite{Arjovsky2017WassersteinGA}, and evolutionary GAN (EGAN) \cite{Wang2019EvolutionaryGA} as baselines for comparison and discussion, because these GAN models are closed to our method.
	\subsubsection{Qualitative Evaluation}\label{sec4.3.1}
\begin{figure*}[htbp]
		\centering
		\includegraphics[width=16cm,height=10cm]{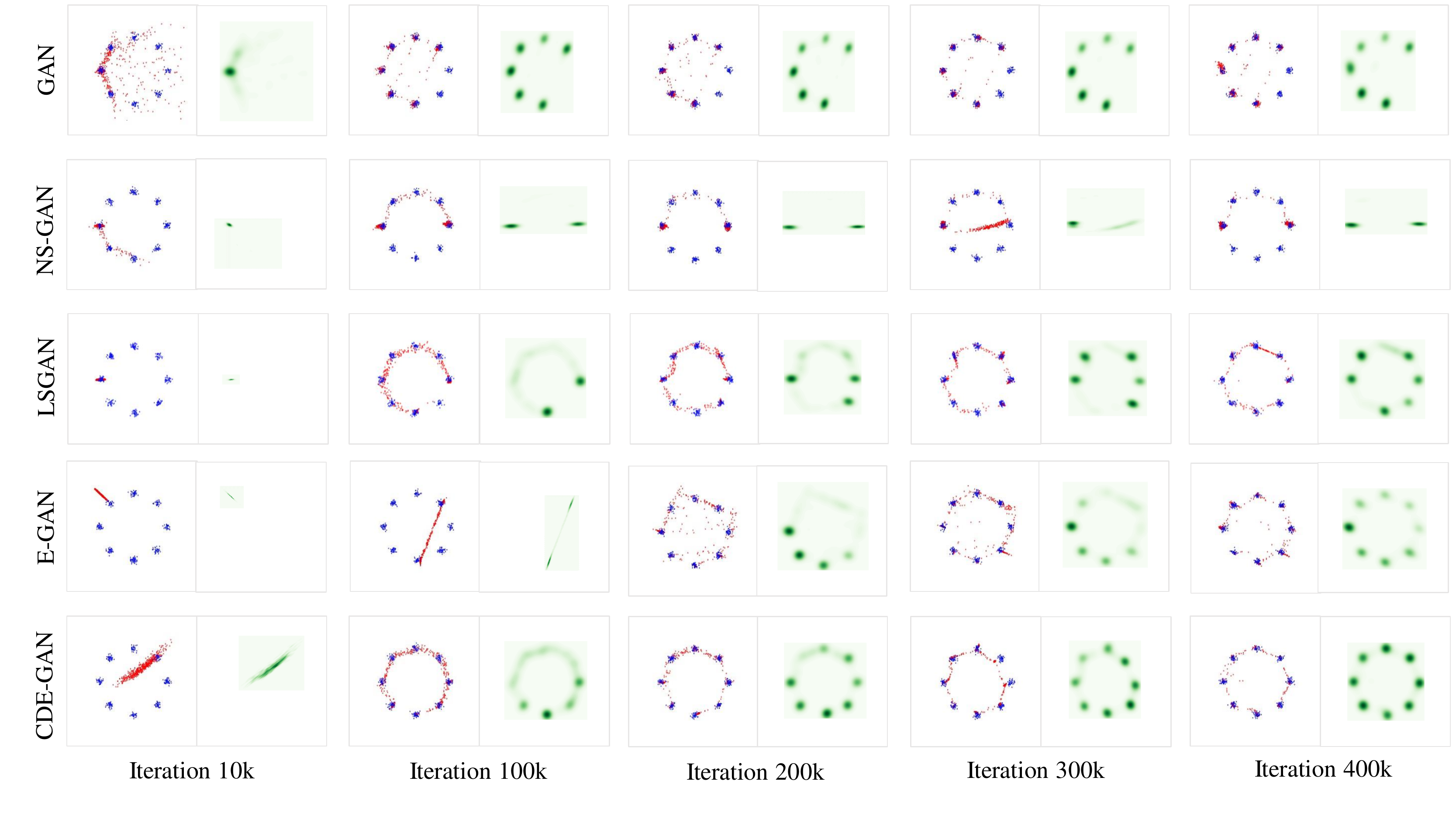}
		\vspace{-3mm}\\(a) Gaussian kernel estimation with MLP of 3 layers\vspace{1mm}
		\includegraphics[width=15.8cm,height=10cm]{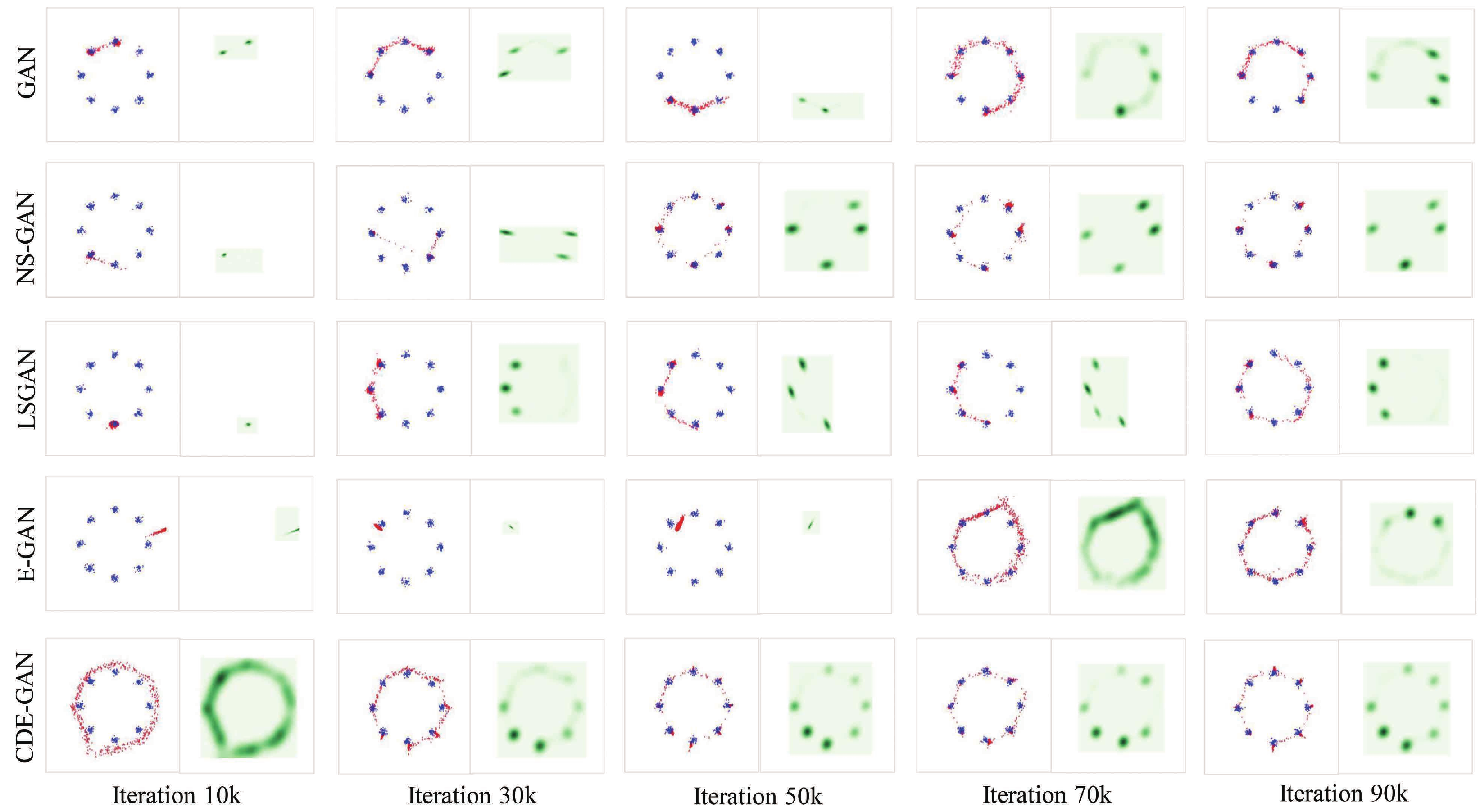}
		\\(b) Gaussian kernel estimation with MLP of 4 layers
		\caption{Dynamic results of Gaussian kernel estimation over generator iteration for different GANs. For each pair of images, the left one is data distribution (real data is represented in blue, generated data is represented in red), and the right one is KDE plots of the generated data corresponded to its left generated data. From top to bottom, the rows are the results of vanilla GAN,  NS-GAN, LSGAN, E-GAN, and CDE-GAN (Ours).}
		\label{fig:difgan-toy}
	\end{figure*}

	Learning on a Gaussian mixture distribution to evaluate the diversity of GANs is a popular experiment setting, which intuitively reveals GANs whether to suffer from mode collapse. When the model suffers from the mode collapse problem, it will generate samples only around a few modes. To validate the effectiveness of our proposed method, analogously, we first qualitatively compare CDE-GAN with different baselines on synthesis dataset of 2-D mixture of 8 gaussian mixture distributions. For conducting a fair comparison, we adopt the experimental design proposed in \cite{Mao2017LeastSG,Mao2019OnTE}, which trains GANs with 3 layers of MLP network architecture. Meanwhile, the survived parents number of E-Discriminators ($I$) and E-Generators ($J$) of CDE-GAN are set as 1, i.e., during each evolutionary step, only the best one candidature in each evolution algorithm is kept. We train each method over 400k generator iterations with the same network architecture. As shown in Fig. \ref{fig:difgan-toy}(a), the dynamic results of data distribution and the Kernel Density Estimation (KDE) plots on different baselines are reported. We can see that all of the baselines only generate samples around a few of valid modes of the data distribution, i.e., 6 modes for vanilla GAN and LSGAN, 2 modes for NS-GAN, 8 modes for E-GAN (parts of modes are weakly covered), which shows that they suffer from mode collapse to a greater or lesser degree. Nevertheless, CDE-GAN can successfully learn the Gaussian mixture distribution for all modes. These experiments demonstrate that the cooperative dual evolutionary strategy well circumvents the mode collapse.
	
	Furthermore, we also try 4 layers of MLP network architecture to conduct the same experiment to evaluate the stability of CDE-GAN. In Fig. \ref{fig:difgan-toy}(b), The results show that all methods can speed up their convergence speed compared to the results on 3 layers MLP architecture. Notably, all baselines tend to generate a few modes, while CDE-GAN is less prone to this problem and achieves better convergence speed. It reveals that CDE-GAN possesses another advantage of architecture robustness.

\begin{figure*}[ht]
		\centering
		\includegraphics[width=5cm,height=5cm]{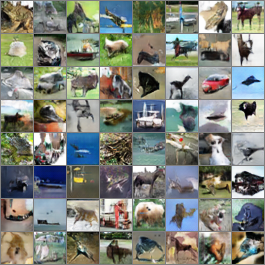}
		\hspace{2.5mm}\includegraphics[width=5cm,height=5cm]{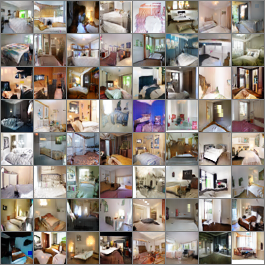}
		\hspace{2.5mm}\includegraphics[width=5cm,height=5cm]{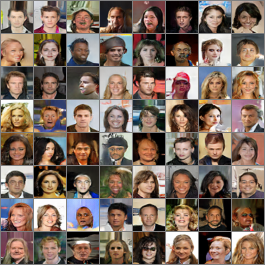}
		\\(a) CIFAR-10 \hspace{2.5cm}  (b) LSUN-Bedrooms \hspace{2.5cm}  (c) CelebA  \\
		\caption{Samples generated by our proposed CDE-GAN on various natural image datasets, i.e., CIFAR-10, LSUN-Bedrooms and CelebA.  Please see many more results in project page.}
		\label{fig:visualization-dif-dataset}
	\end{figure*}

	For the sake of proving the potential power of generative performance of CDE-GAN, we show several samples generated by our proposed model trained on three datasets with $32\times32$ pixels in Fig. \ref{fig:visualization-dif-dataset}, i.e., CIFAR-10, LSUN-Bedrooms, and CelebA. Note that the presented samples are fair random drew, not cherry-picked. It can be seen, on CIFAR-10, that CDE-GAN is capable of generating visually recognizable images of frogs, airplanes, horses, etc. CDE-GAN can produce bedrooms with various styles and views, and beds and windows in the rooms are clearly displayed on LSUN-Bedrooms. It can also synthesize face images possessing various attributes (e.g., gender, age, expression, and hairstyle). These results confirm the quality and diversity of samples generated by our method.

	\begin{figure}[t]
		\centering
		\includegraphics[width=1\linewidth]{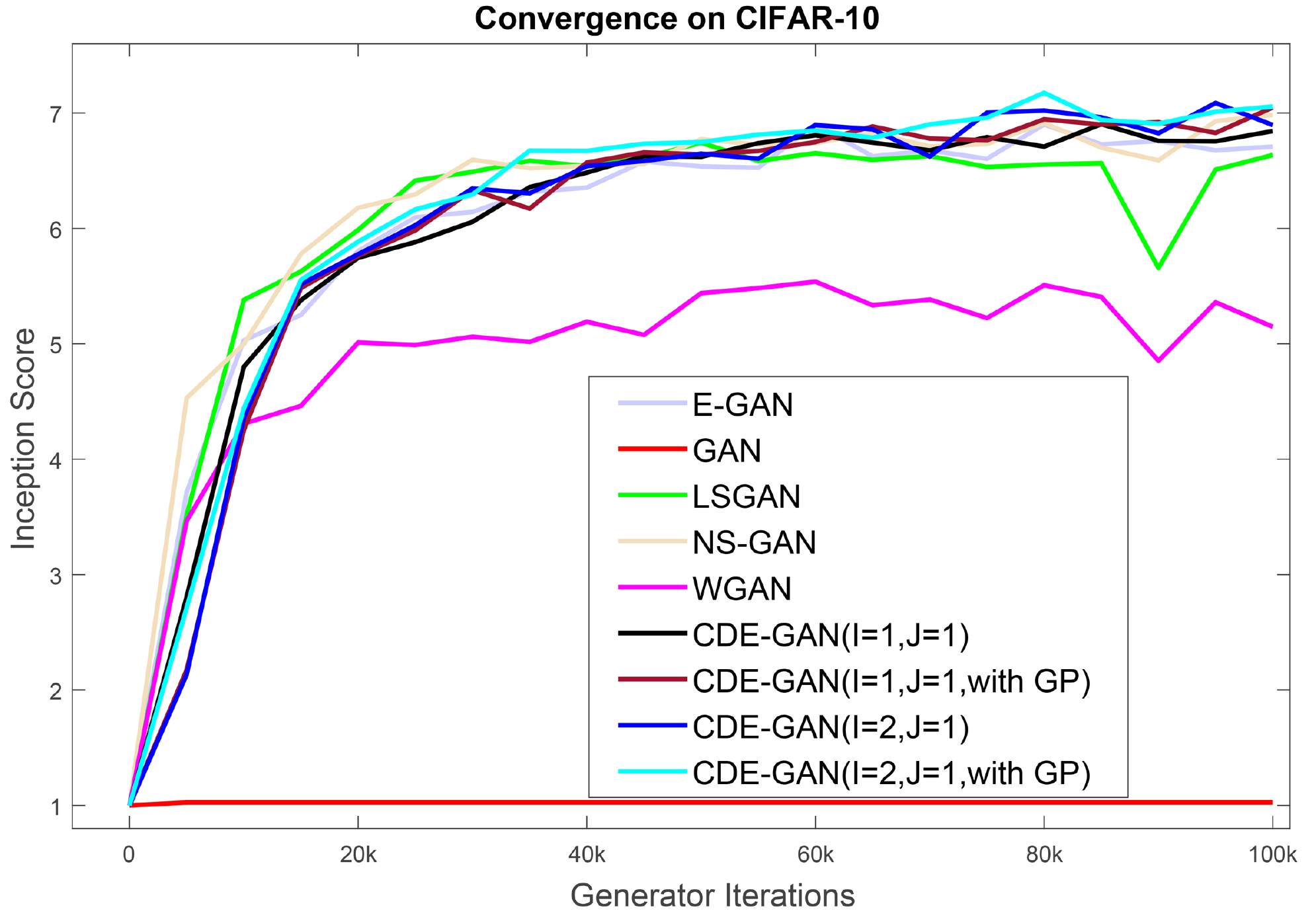}
		\caption{Inception score of different GANs on CIFAR-10.}
		\label{fig:IS-CIFAR10}
	\end{figure}

	\subsubsection{Quantitative Evaluation}\label{sec4.3.2}
	Our qualitative observations above are confirmed by the quantitative evaluations. For the sake of demonstrating the merits of the proposed CDE-GAN over the baselines, we train these methods on CIFAR-10 and plot inception scores over the training process with the same network architecture. As shown in Fig. \ref{fig:IS-CIFAR10}, CDE-GAN can get a higher inception score within 100k generator iterations. After 40k iterations, specifically, CDE-GANs with different settings consistently perform better over all the baselines. Meanwhile, the baselines fall in different training problems, e.g., instability at convergence (wgan and lsgan), invalid (GAN). This shows that the cooperative dual evolution method is benefiting for adversarial training of GANs. 
	
	\begin{table}[t]
		\centering
		\captionsetup{justification=centering}
		\caption{\textsc{Comparison with E-GAN on CIFAR-10 With or Without GP.} $\dagger$\cite{Wang2019EvolutionaryGA} }\label{Table:comparison-EGAN}
		\setlength{\tabcolsep}{4mm}{
			\begin{tabular}{l|c}
				\hline       
				\textbf{Methods}              & \textbf{Inception Score} \\   
				\hline
				E-GAN ($\mu=1$, without GP)   & 6.88 $\pm$ 0.10  \\
				E-GAN ($\mu=2$, without GP)   & 6.71 $\pm$ 0.06  \\
				E-GAN ($\mu=4$, without GP)   & 6.96 $\pm$ 0.09   \\
				E-GAN ($\mu=8$, without GP)   & 6.72 $\pm$ 0.09  \\
				E-GAN ($\mu=1$, with GP)$\dagger$      & 7.13 $\pm$ 0.07  \\
				E-GAN ($\mu=2$, with GP)$\dagger$      & 7.23 $\pm$ 0.08  \\
				E-GAN ($\mu=4$, with GP)$\dagger$      & 7.32 $\pm$ 0.09 \\
				E-GAN ($\mu=8$, with GP)$\dagger$      & 7.34 $\pm$ 0.07  \\
				\hline
				(Ours) CDE-GAN ($I=1$, without GP) & 6.85 $\pm$ 0.05  \\
				(Ours) CDE-GAN ($I=2$, without GP) & 6.93 $\pm$ 0.09  \\
				(Ours) CDE-GAN ($I=4$, without GP) & 7.06 $\pm$ 0.09  \\
				(Ours) CDE-GAN ($I=8$, without GP) & 7.35 $\pm$ 0.06  \\
				(Ours) CDE-GAN ($I=1$, with GP)    & 7.05 $\pm$ 0.05  \\
				(Ours) CDE-GAN ($I=2$, with GP)    & 7.18 $\pm$ 0.05  \\
				(Ours) CDE-GAN ($I=4$, with GP)    & 7.48 $\pm$ 0.10  \\
				(Ours) CDE-GAN ($I=8$, with GP)    & \textbf{7.51 $\pm$ 0.05}  \\
				\hline
				
		\end{tabular}}
	\end{table}

	Since EGAN \cite{Wang2019EvolutionaryGA} is most similar to our method, we further take E-GAN as a baseline to compare with CDE-GAN for stability analysis. In Table \ref{Table:comparison-EGAN}, we take various number of generators $\mu=\{1,2,4,8\}$ for E-GAN and various number of discriminators $I=\{1,2,4,8\}$ for CDE-GAN. We train each model in 150k generator iterations to conduct experiments. Note that we implement results of E-GAN using the experimental setting of literature \cite{Wang2019EvolutionaryGA} and codes provided by authors\footnote{https://github.com/WANG-Chaoyue/EvolutionaryGAN-pytorch}.  Results show that CDE-GAN achieves better performance on CIFAR-10 than E-GAN. Furthermore, if E-GAN uses GP term during training, its generative performance is greatly improved compared to the results of its original version (e.g., 0.62 improvement for IS when $\mu$ is set as 8); while our CDE-GAN achieves a little improvement if GP is used (e.g., 0.16 improvement for IS when $I$ is set as 8). This reveals that E-GAN is unstable during training due to its single evolution, and thus GP term is effective for regularizing the discriminator to provide informative gradients for updating the generators. Benefiting from cooperative dual evolution, CDE-GAN injects dual diversity into training, and thus it can cover different data modes. Furthermore, with the number of discriminator increasing, the balance of generator and discriminators of CDE-GAN is well adjusted and thus it continuously achieves obvious improvement. This shows that adversarial multi-objective optimization is effective for stabilizing the training process of CDE-GAN. To this end, the discriminators can continuously provide informative gradient for generator's updation, which performs the function of GP.
	
	\begin{figure*}[htbp]
		\centering
		\includegraphics[width=5.7cm,height=5.7cm]{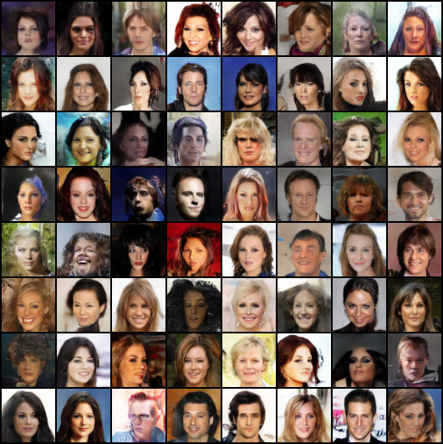}
		\hspace{2.5mm}\includegraphics[width=5.7cm,height=5.7cm]{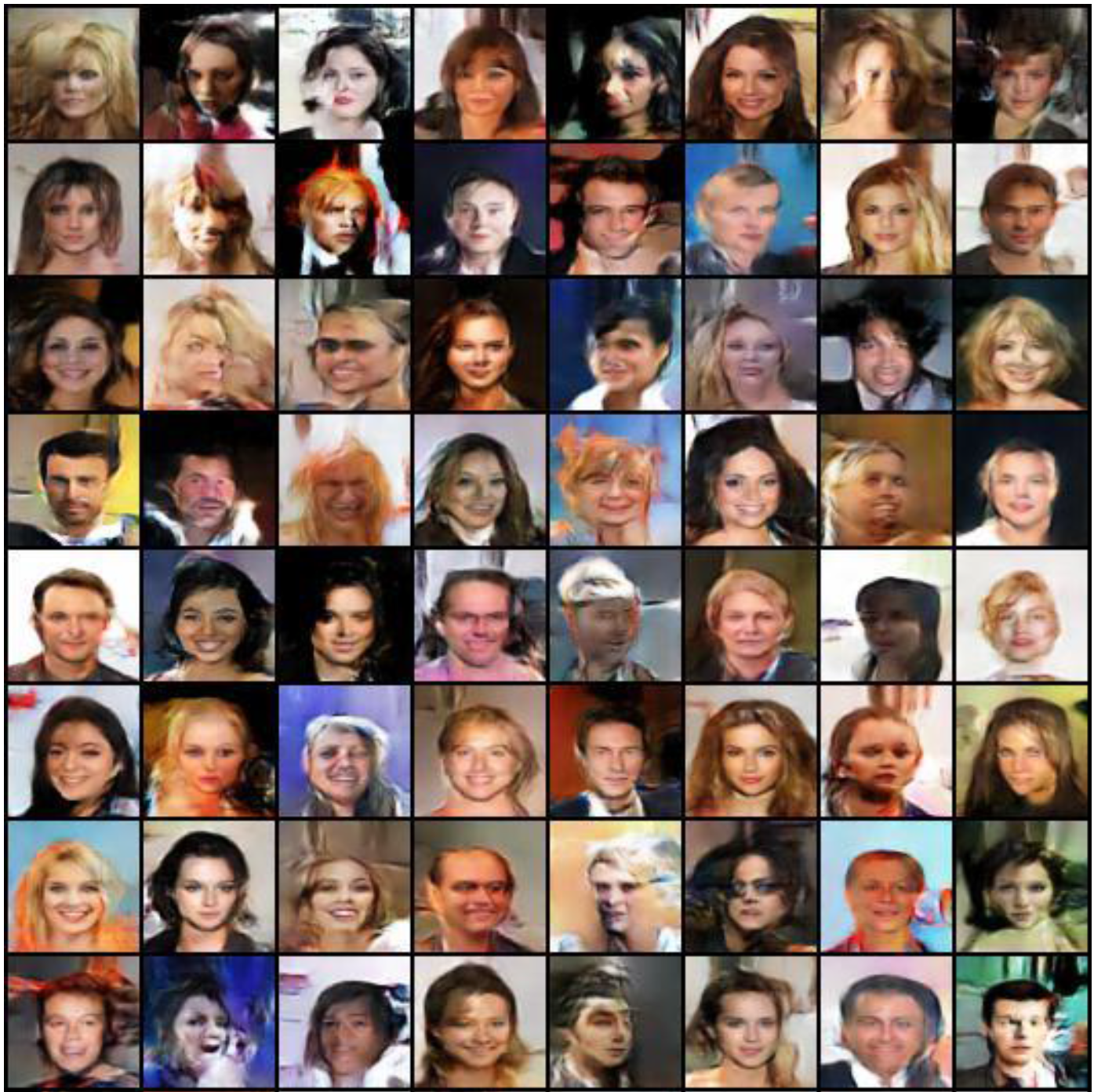}
		\hspace{2.5mm}\includegraphics[width=5.7cm,height=5.7cm]{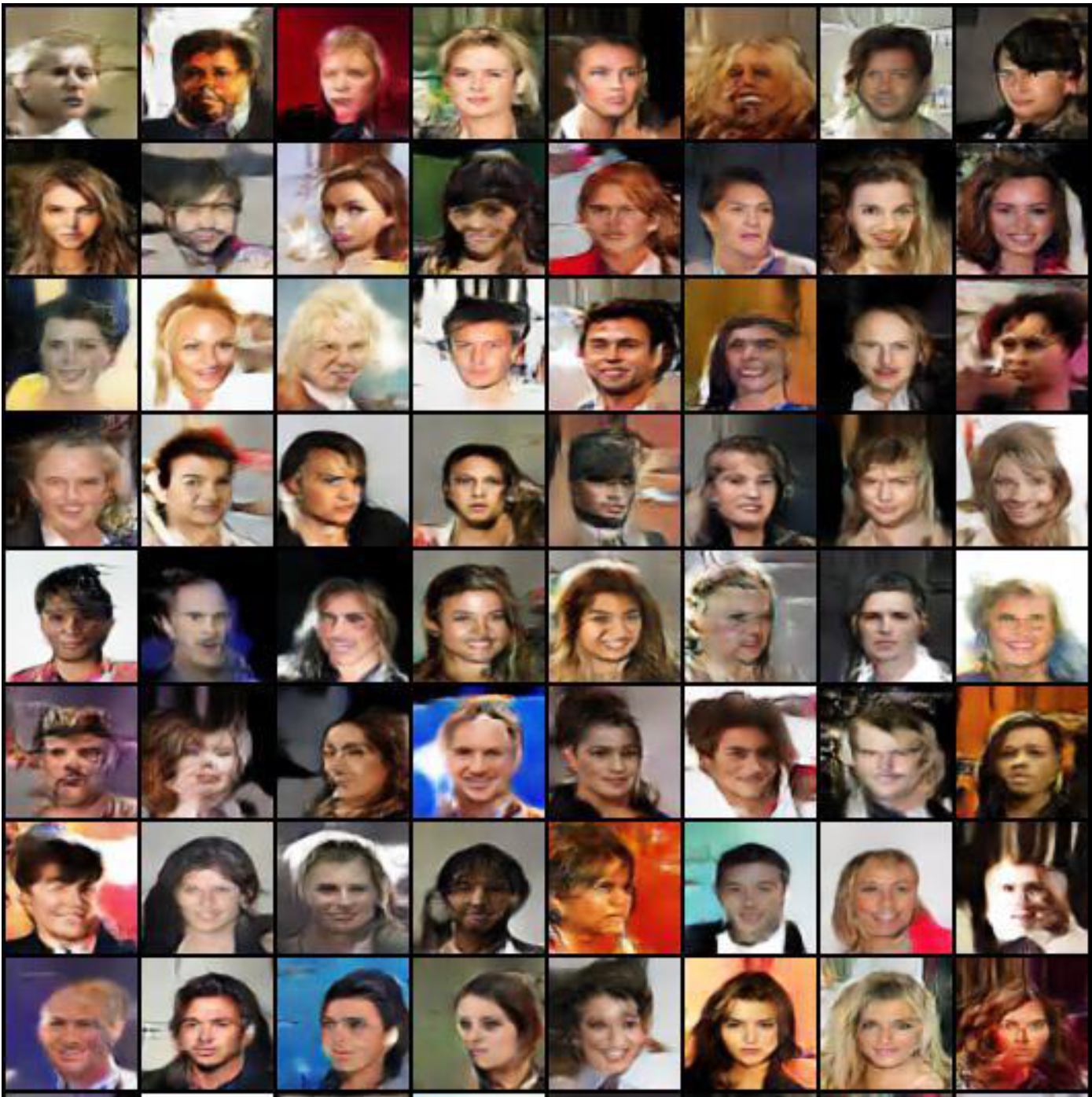}
		\\ \hspace{-0.2cm}(a) MAD-GAN($k=3$) \hspace{3cm}  (b)  Lipizzaner \hspace{3.5cm}  (c)  Mustangs  \\ \vspace{1.5mm}
		\includegraphics[width=5.7cm,height=5.7cm]{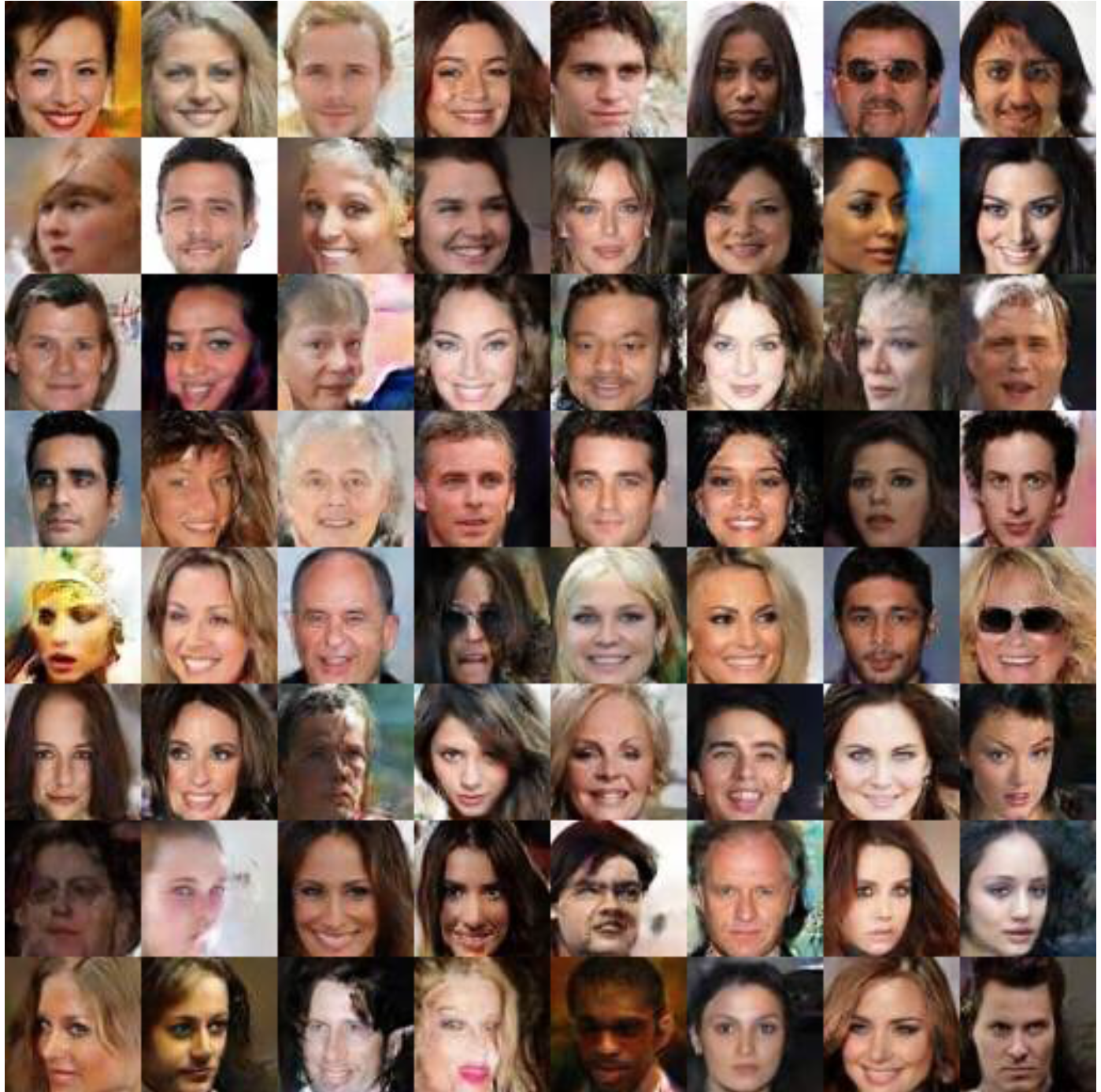}
		\hspace{2.5mm}\includegraphics[width=5.7cm,height=5.7cm]{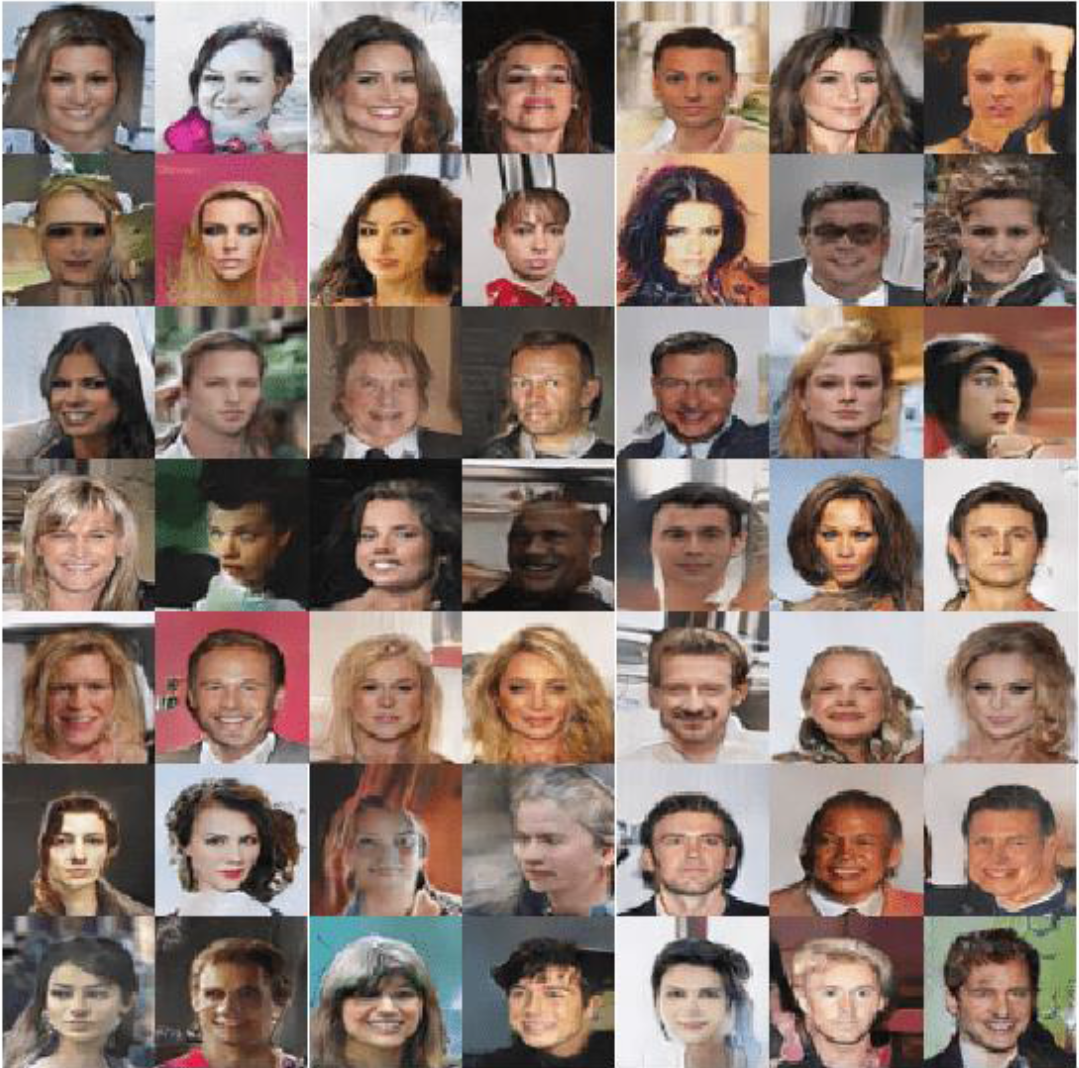}
		\hspace{2.5mm}\includegraphics[width=5.7cm,height=5.7cm]{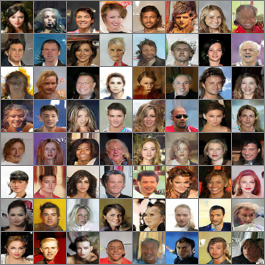}
		\\ \hspace{-0.2cm}(d) Stabilizing-GAN($K=24$) \hspace{2cm}  (e) acGAN($N=5$)   \hspace{2.2cm}  (f) \textbf{CDE-GAN($I=2$)}   \\
		\caption{Visual comparisons of different methods on CelebA. The samples generated by different methods are provided by the original literatures, i.e., MAD-GAN \cite{Ghosh2018MultiagentDG}, Lipizzaner \cite{Schmiedlechner2018TowardsDC}, Mustangs \cite{Toutouh2019SpatialEG}, Stabilizing-GAN \cite{Neyshabur2017StabilizingGT}, and acGAN \cite{Doan2019OnlineAC}.}
		\label{fig:visualization-comparison}
	\end{figure*}

	\begin{table}[htbp]
		\centering
		\captionsetup{justification=centering}
		\caption{\textsc{Comparison with State-of-the-Art Methods on CIFAR-10. The $N$, $K$, $T$, $\mu$, and $I$ Represent the Numbers of Discriminators or Generators for Different Methods, \textit{Addit. Superv. Info.} Denotes That Additional Supervised Information is Used by Mehtod. The Best Two Results are Marked with \textbf{Bold}, and \underline{Underline}.}}\label{Table:comparison-dif-gan-CIFAR}
		\setlength{\tabcolsep}{2mm}{
			\begin{tabular}{l|c|c}
				\hline       
				\textbf{Methods}  & \textbf{Addit. Superv. Info.}             & \textbf{Inception Score} \\   
				\hline
				Real data        &             & 11.24$\pm$ 0.12 \\
				\hline
				GAN-GP  \cite{Goodfellow2014GenerativeAN}              &  \XSolidBrush        & 6.93 $\pm$ 0.08  \\
				DCGAN \cite{Radford2015UnsupervisedRL}                 &  \XSolidBrush        & 6.64 $\pm$ 0.14  \\
				WGAN-GP  \cite{Gulrajani2017ImprovedTO}              &  \XSolidBrush        & 6.68 $\pm$ 0.06  \\
				SN-GAN \cite{Miyato2018SpectralNF}                     &  \XSolidBrush        & 7.42 $\pm$ 0.08  \\
				\hline
				GMAN ($N=5$) \cite{Durugkar2017GenerativeMN}           &  \XSolidBrush        & 6.40 $\pm$ 0.19  \\
				D2GAN \cite{Nguyen2017DualDG}                          &  \XSolidBrush        & 7.15 $\pm$ 0.07  \\
				HV($K=24$) \cite{Albuquerque2019MultiobjectiveTO}      &  \XSolidBrush        & 7.32 $\pm$ 0.26  \\
				MicroGAN ($K=2$)\cite{Mordido2020microbatchGANSD}      &  \XSolidBrush        & 6.77 $\pm$ 0.00  \\
				\hline
				MIX+WGAN ($T=5$) \cite{Arora2017GeneralizationAE}      &  \XSolidBrush        & 4.36 $\pm$ 0.04 \\
				MGAN ($K=10$) \cite{Hoang2018MGANTG}                   &  \Checkmark          & \textbf{8.33 $\pm$ 0.10} \\
				\hline
				E-GAN ($\mu=8$) \cite{Wang2019EvolutionaryGA}          &  \XSolidBrush        & 7.34 $\pm$ 0.07 \\
				(Ours) CDE-GAN ($I=8$)                                 &  \XSolidBrush        & \underline{7.51 $\pm$ 0.05} \\
				\hline
				
		\end{tabular}}
	\end{table}
	
	\subsection{Experiment 3: Comparisons With State-of-the-Art Methods}\label{sec4.4}
	In this section, we compare CDE-GAN with 11 state-of-the-art GANs to show its advantages. We first report the IS obtained by our CDE-GAN and baselines on CIFAR-10 in Table \ref{Table:comparison-dif-gan-CIFAR}. Overall, experimental results show that CDE-GAN outperforms almost all baselines on the optimal settings (except MGAN).  The reason is that MGAN takes \textit{additional supervised information} (generators' labels) for supporting semi-supervised learning. Nevertherless, CDE-GAN is an entirely unsupervised manner, this is the concentration of this paper. Thus, integrating it into evolutionary GANs to conduct semi-supervised learning could be a promising avenue for our future work.
	
	It is worthy to note that CDE-GAN terms adversarial training as a multi-objective optimization problem when $I\ge 2$. Thus we further discuss the advantage of CDE-GAN compared with multi-discriminator based GANs. In the second group of Table \ref{Table:comparison-dif-gan-CIFAR}, all multi-discriminator based GANs using various numbers of discriminators are inferior to our CDE-GAN, i.e., GMAN ($K=5$), D2GAN (2 discriminators), HV($K=24$), MicroGAN ($K=2$). This further demonstrates that cooperative dual evolution is effective for multi-objective optimization and adversarial training of GANs.
	
	Finally, we display visual quality comparisons between several state-of-the-art methods (i.e., Stabilizing-GAN \cite{Neyshabur2017StabilizingGT}, acGAN \cite{Doan2019OnlineAC}, MAD-GAN \cite{Ghosh2018MultiagentDG}, Lipizzaner \cite{Schmiedlechner2018TowardsDC}, and Mustangs \cite{Toutouh2019SpatialEG}) and our method on CelebA. See from  Fig. \ref{fig:visualization-comparison}, the most of the face images generated by Lipizzaner and Mustangs are incomplete or blurred, while our CDE-GAN generates high-fidelity face images. This demonstrates that the proposed cooperative dual evolutionary strategy performs better on adversarial training compared to other evolutionary computation based GANs. Furthermore, CDE-GAN also performs advantages over other GAN methods, i.e., MAD-GAN with 3 generators, Stabilizing-GAN with 24 discriminators, acGAN with 5 discriminators. Note that MAD-GAN trained its model using \textit{additional supervised information} (generators' labels) and Stabilizing-GAN used the cropped version of the images of CelebA. More visual comparisons on CIFAR-10 and LSUN-Bedrooms are provided in the supplementary file.

	\section{Discussion}\label{sec5}
	
	We would like to have more discussion here about the advantages and limitations of the proposed CDE-GAN method. First, we analyze why and how  our method can circumvent the mode collapse and instability problem of GANs.
	\begin{itemize}
		
		\item  Compared to single evolution strategy (i.e., E-GAN \cite{Wang2019EvolutionaryGA}), CDE-GAN injects dual diversity into training benefiting from the cooperative dual evolution. It decomposes the complex adversarial optimization problem into two subproblems (i.e., generation and discrimination), and each subproblem is solved by a separated subpopulation (i.e., E-Generators and E-Discriminators), evolved by its own evolutionary algorithm (including individual variations, evaluation, and selection). Furthermore, The complementary mutations in E-Generators and E-Discriminators are helpful for CDE-GAN to evolve in different possible directions during various training stages. In this way, the dual evolutionary population injects dual diversity into the training, and thus it can effectively cover different data modes. This significantly mitigates training pathologies of mode collapse for GANs. Experiments in Section \ref{sec4.3} and \ref{sec4.4} intuitively verify this claims. 
		
		\item  CDE-GAN terms adversarial training as an adversarial multi-objective optimization problem, and thus the multiple discriminators provide informative feedback gradient to the generator for stabilizing training process. In ideal, we prefer that generator always has strong gradients from the discriminator during training. Since the discriminator quickly learns to distinguish real and fake samples, the single-objective optimization-based GANs make this difficult to ensure. To this end, they cannot provide meaningful error signals to improve the generator thereafter. In contrast, multi-objective optimization simultaneously optimizes the losses provided by different models to favor worse discriminators. Thus discriminators provide more informative gradients to the generator. The experiments in Section \ref{sec4.2.2} and \ref{sec4.3.2} support this conclusion.
		
		\item Soft Mechanism well balances the trade-off between E-Generators and E-Discriminators to help CDE-GAN conduct effective adversarial training. In fact, the degenerate results of GANs can be avoided by employing learner (discriminator) with limited capacity and corrupting data samples with noise \cite{Neyshabur2017StabilizingGT,Durugkar2017GenerativeMN,Doan2019OnlineAC}. To this end, Soft Mechanism softens the maximization of discriminators to weaken the discriminators, which avoids the problem of whack-a-mole dilemma during training and enables stable training of CDE-GAN. Section \ref{sec3.1} provides more theoretical analysis.

	\end{itemize}

	Indeed, CDE-GAN limits in costing more time in each iteration. Theoretically, let's take updating a generator iteration as a training step. CDE-GAN will cost $\mathcal{O}(K \cdot I\cdot N)+\mathcal{O}(J \cdot M)$ operations for one step (all notations are defined in Algorithm \ref{algotithm:CDE-GAN}, and $J$ is set as 1 in this paper). In practice, the time-consuming of CDE-GANs at each iteration is reported in Table \ref{Table:time-consuming}. Specifically, to train a CDE-GAN model for $32\times32$ images using DCGAN architecture with different numbers of discriminators ($I=\{1, 2, 4, 8\}$), it will cost around $0.079\pm 0.003$, $0.079\pm 0.003$, $0.158\pm 0.021$, and $0.303\pm 0.002$ seconds respectively for one generator iteration (excluding generating images for score test) on a single GPU. Since the cooperative dual evolutionary strategy is effective for adversarial training, CDE-GAN performs significantly fewer training steps to achieve the same generative performance than other baselines (see Fig. \ref{fig:IS-CIFAR10}). 
	

	\begin{table}[hbpt]
		\centering
		\captionsetup{justification=centering}
		\caption{\textsc{Time Consuming of CDE-GANs at Each Generator Iteration.}}\label{Table:time-consuming}
		\setlength{\tabcolsep}{8mm}{
			\begin{tabular}{l|c}
				\hline       
				\textbf{Methods}              & \textbf{Time/Iteration (Seconds)} \\   
				\hline
				CDE-GAN($I=1$)                    & 0.039$\pm$ 0.000 \\
				CDE-GAN($I=2$)                    & 0.079$\pm$ 0.003 \\
				CDE-GAN($I=4$)                    & 0.158$\pm$ 0.021 \\
				CDE-GAN($I=8$)                    & 0.303$\pm$ 0.002 \\
				\hline
				
		\end{tabular}}
	\end{table}

	\section{Conclusion}\label{sec6}
	In this paper, we proposed a novel GAN (CDE-GAN), incorporating cooperative dual evolution with respect to E-Generators and E-Discriminators into a unified evolutionary adversarial framework, to circumvent adversarial optimization difficulties of GANs, i.e., mode collapse and instability. Notably, the dual evolution provides a dynamic strategy to the generator(s) and discriminators, exploits the complementary properties, and injects dual mutation diversity into learning. This diversifies the estimated density in capturing multi-modes and improves the generative performance of CDE-GAN. Additionally, we introduced a Soft Mechanism to balance E-Generators and E-Discriminators for conducting effective and stable training. The competitive results on one synthetic dataset (i.e., the 2D mixture of 8 Gaussian) and three real-world image datasets (i.e., CIFAR-10, LSUN-Bedrooms, and CelebA), demonstrate the superiority and great potentials of cooperative dual evolution for GANs. Extensive experiments also show that CDE-GAN performs obvious advantages over all the compared state-of-the-art methods.

	In the future, we will further improve CDE-GAN for incorporating additional supervised information into learning and speeding up learning. Meanwhile, we will also apply evolutionary computation based GANs to other generative tasks, e.g., text synthesis, video prediction. 
	
	\section{ACKNOWLEDGMENT}\label{sec7}
	The authors would like to thank Dr. Chaoyue Wang at the School of Computer Science, University of Sydney, for his assistance with coding and theoretical suggestions.

	
	%

	



	\ifCLASSOPTIONcaptionsoff
	\newpage
	\fi

	
	
	%
	
	\bibliographystyle{IEEEtran}
	\bibliography{mybibfile}
	
	
	
	
	%
	
	
	\vspace{-1.35cm}
	\begin{IEEEbiography}[{\includegraphics[width=1in,height=1.25in,clip,keepaspectratio]{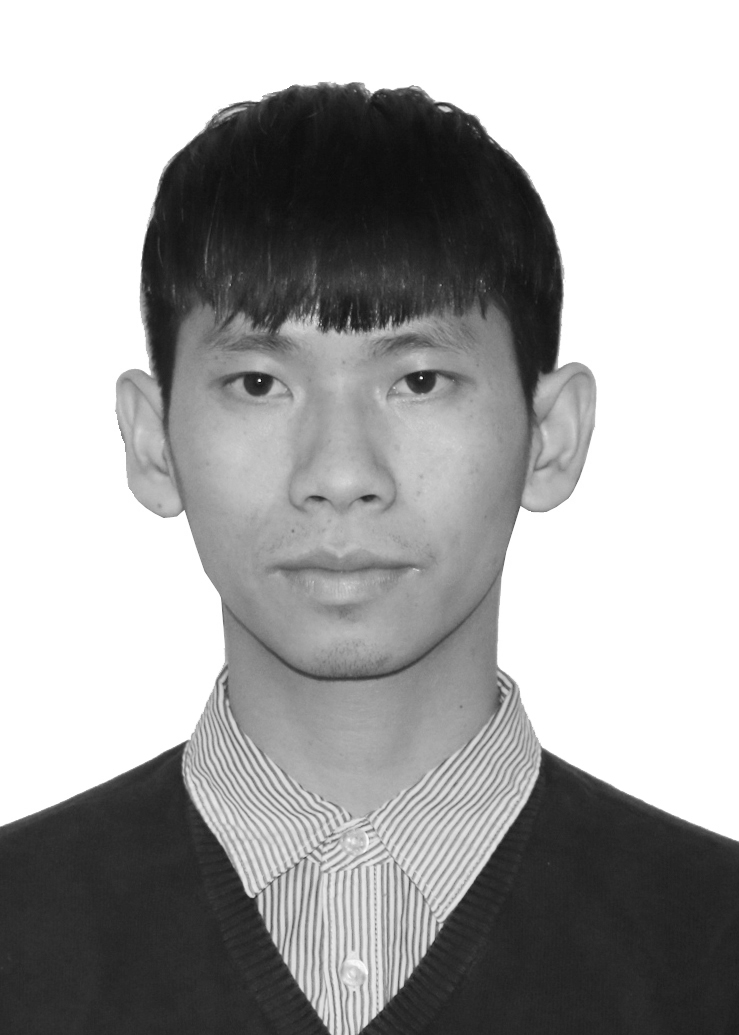}}]{Shiming Chen}
		is currently a full-time Ph.D student in the School of Electronic Information and Communications, Huazhong University of Sciences and Technology (HUST), China. He serves as the reviewer for prestigious journals such as \textit{IEEE Transactions on Image Processing, IEEE Transactions on Systems, Man, and Cybernetics: Systems, IEEE Transactions on Industrial Informatics, Information Fusion, Information Sciences, and Applied Soft Computing}. His current research interests span computer vision and machine learning with a series of topics, such as generative modeling and learning, domain adaptation, and zero-shot learning. 
	\end{IEEEbiography}
	
	\vspace{-1.1cm}
	\begin{IEEEbiography}[{\includegraphics[width=1in,height=1.25in,clip,keepaspectratio]{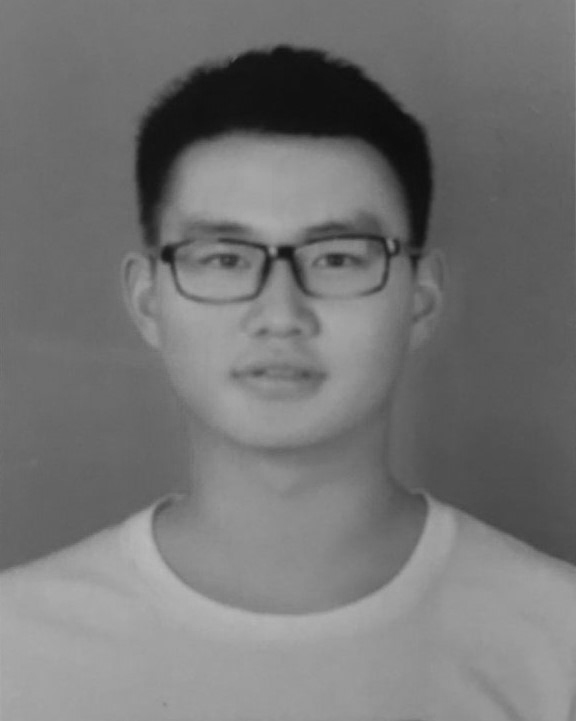}}]{Wenjie Wang}
		is currently pursuing the M.Sc. degree in the School of Electronic Information and Communications(EIC), Huazhong University of Sciences and Technology(HUST), China, in 2019. He received the B.E. degree in the School of EIC, HUST. His current research interests include multimodal learning, computer vision, and machine learning.
	\end{IEEEbiography}
	
	\vspace{-1.2cm}
	\begin{IEEEbiography}[{\includegraphics[width=1in,height=1.25in,clip,keepaspectratio]{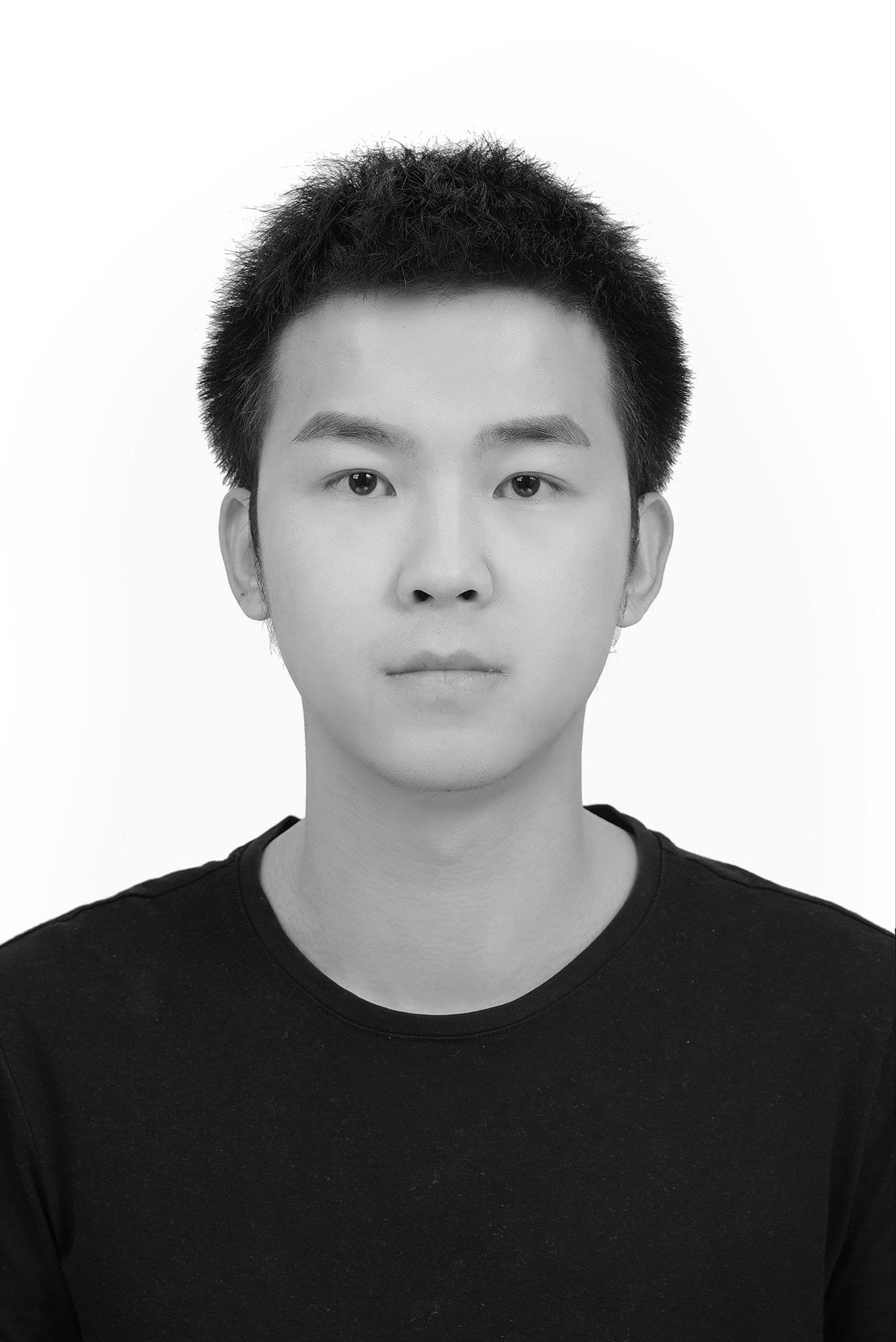}}]{Beihao Xia}
		is currently pursuing the Ph.D. degree in the School of Electronic Information and Communications (EIC), Huazhong University of Sciences and Technology(HUST), China. He received the B.E. degree in College of Computer Science and Electronic Engineering, Hunan University(HNU), China, in 2015, and the M.Sc. degree in the School of EIC, HUST, China, in 2018, respectively. His current research interests include image/video processing, computer vision, and machine learning.
	\end{IEEEbiography}   
	
	\begin{IEEEbiography}[{\includegraphics[width=1in,height=1.25in,clip,keepaspectratio]{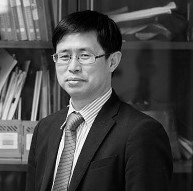}}]{Xinge You}
		(Senior Member, IEEE) is currently a Professor with the School of Electronic Information and Communications, Huazhong University of Science and Technology, Wuhan. He received the B.S. and M.S. degrees in mathematics from Hubei University, Wuhan, China, in 1990 and 2000, respectively, and the Ph.D. degree from the Department of Computer Science, Hong Kong Baptist University, Hong Kong, in 2004. His research results have expounded in 20+ publications at prestigious journals and prominent conferences, such as  IEEE T-PAMI, T-IP, T-NNLS, T-CYB, T-CSVT, CVPR, ECCV, IJCAI. He served/serves as an Associate Editor of the \textit{IEEE Transactions on Cybernetics}, \textit{IEEE Transactions on Systems, Man, Cybernetics:Systems}. His current research interests include image processing, wavelet analysis and its applications, pattern recognition, machine earning, and computer vision.
	\end{IEEEbiography}
	
	\vspace{-0.6cm}
	\begin{IEEEbiography}[{\includegraphics[width=1in,height=1.25in,clip,keepaspectratio]{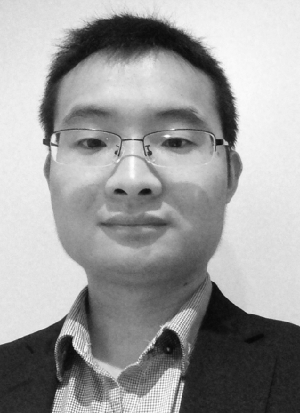}}]{Qinmu Peng}
		is currently an Assistant Professor in the School of Electronics Information and Communications, Huazhong University of Science and Technology (HUST), China. He received the Ph.D. degree in computer science from Hong Kong Baptist University, Hong Kong, in 2015. His research results have expounded in 20+ publications at prestigious journals and prominent conferences, such as PNAS, IEEE T-NNLS, T-SMCA, T-HMS, T-MM,  IJCAI. His current research interests include multimedia analysis, computer vision, and medical image analysis.
	\end{IEEEbiography}
	
	\vspace{-0.6cm}
	\begin{IEEEbiography}[{\includegraphics[width=1in,height=1.25in,clip,keepaspectratio]{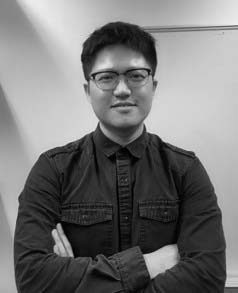}}]{Zehong Cao}
		is a Lecturer with the Discipline
		of Information and Communication Technology, School of Technology,
		Environments and Design, University of Tasmania, Australia, and an Adjunct Fellow with the School of Computer Science, UTS. He received the Ph.D. degree in information technology from the University of Technology Sydney, Australia, in 2017.He has authored more than 50 papers published in top-tier International Conferences such as AAMAS and AAAI, and IEEE and ACM Transactions Series, with 5 ESI highly cited papers. Dr. Cao is the Leading Guest Editor of \textit{IEEE Transactions on Fuzzy Systems}, and \textit{IEEE Transactions on Industrial Informatics} (2020), and the Associate Editor for \textit{Neurocomputing} (2019-) and \textit{Scientific Data} (2019-). His current research interests include computer vision, machine learning, computational intelligence.
	\end{IEEEbiography}

	\vspace{-0.6cm}
	\begin{IEEEbiography}[{\includegraphics[width=1in,height=1.25in,clip,keepaspectratio]{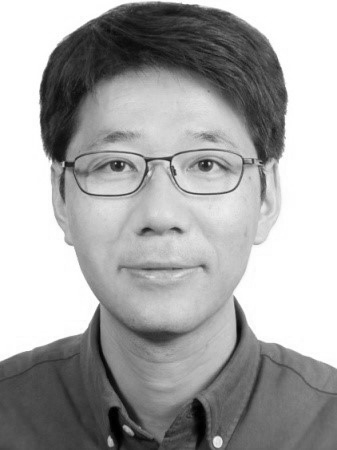}}]{Weiping Ding}
		(M'16-SM'19) is currently a Full Professor with the School of Information Science and Technology, Nantong University, Nantong, China. He received the Ph.D. degree in Computation Application, Nanjing University of Aeronautics and Astronautics (NUAA), Nanjing, China, in 2013. He was a Visiting Scholar at the University of Lethbridge(UL), Alberta, Canada, in 2011. From 2014 to 2015, He is a Postdoctoral Researcher at the Brain Research Center, National Chiao Tung University (NCTU), Hsinchu, Taiwan. In 2016, He was a Visiting Scholar at the National University of Singapore (NUS), Singapore. From 2017 to 2018, he was a Visiting Professor at the University of Technology Sydney (UTS), Ultimo, NSW, Australia. He has published more than 80 research peer-reviewed journal and conference papers, including IEEE T-FS, T-NNLS, T-CYB, T-BME, T-II, T-ETCI, T-ITS and CIKM, etc. He has three ESI highly cited papers. Dr. Ding currently serves on the Editorial Advisory Board of \textit{Knowledge-Based Systems} and Editorial Board of \textit{Information Fusion}, \textit{Applied Soft Computing}. He serves/served as an Associate Editor of \textit{IEEE Transactions on Fuzzy Systems}, \textit{Information Sciences}, \textit{Swarm and Evolutionary Computation}, and \textit{Journal of Intelligent \& Fuzzy Systems}, and Co-Editor-in-Chief of \textit{Journal of Artificial Intelligence and System}. He is also the Leading Guest Editor of \textit{IEEE Transactions on Evolutionary Computation}  and \textit{Information Fusion} (2020). His main research directions involve data mining, granular computing, evolutionary computing, machine learning, and big data analytics.
	\end{IEEEbiography}
	

\end{document}